\documentclass{article}
\usepackage{ijcai25}

\usepackage{times}
\usepackage{soul}
\usepackage{url}
\usepackage[hidelinks]{hyperref}
\usepackage[utf8]{inputenc}
\usepackage[small]{caption}
\usepackage{graphicx}
\usepackage{amsmath}
\usepackage{amsthm}
\usepackage{booktabs}
\usepackage{algorithm}
\usepackage{algorithmic}
\usepackage[switch]{lineno}

\usepackage{amssymb}
\usepackage{multirow}
\usepackage{xcolor}
\usepackage{subfigure}
\usepackage{hyperref}
\usepackage{colortbl}

\newcommand{\eg}{\textit{e.g.}}
\newcommand{\ie}{\textit{i.e.}}
\newcommand{\wrt}{\textit{w.r.t.}}
\newcommand{\etal}{\textit{et al.}}
\newcommand{\aka}{\textit{a.k.a.}}

\renewcommand{\b}{\color{blue}}

\title{Temporal Consistency Constrained Transferable Adversarial Attacks with Background Mixup for Action Recognition}

\author{
Ping Li$^1$
\and
Jianan Ni$^1$\and
Bo Pang$^{1}$\\
\affiliations
$^1$School of Computer Science and Technology, Hangzhou Dianzi University\\
\emails
 \{lpcs, njn, pbcs\}@hdu.edu.cn
}

\begin{document}

\maketitle

\begin{abstract}
Action recognition models using deep learning are vulnerable to adversarial examples, which are transferable across other models trained on the same data modality. Existing transferable attack methods face two major challenges: 1) they heavily rely on the assumption that the decision boundaries of the surrogate (\aka, source) model and the target model are similar, which limits the adversarial transferability; and 2) their decision boundary difference makes the attack direction uncertain, which may result in the gradient oscillation, weakening the adversarial attack. This motivates us to propose a \textbf{B}ackground \textbf{M}ixup-induced \textbf{T}emporal \textbf{C}onsistency (\textbf{BMTC}) attack method for action recognition. From the input transformation perspective, we design a model-agnostic background adversarial mixup module to reduce the surrogate-target model dependency. In particular, we randomly sample one video from each category and make its background frame, while selecting the background frame with the top attack ability for mixup with the clean frame by reinforcement learning. Moreover, to ensure an explicit attack direction, we leverage the background category as guidance for updating the gradient of adversarial example, and design a temporal gradient consistency loss, which strengthens the stability of the attack direction on subsequent frames. Empirical studies on two video datasets, \ie, \textit{UCF101} and \textit{Kinetics-400}, and one image dataset, \ie, \textit{ImageNet}, demonstrate that our method significantly boosts the transferability of adversarial examples across several action/image recognition models.   
\end{abstract}

\section{Introduction}
Action recognition has established itself as a fundamental task in computer vision, and has widespread applications in many areas, such as video surveillance, robot, and virtual reality. In recent years, Deep Neural Networks (DNNs) have gained large popularity in developing action recognition models, such as SlowFast \cite{feichtenhofer-iccv2019-slowfast} and Video Vision Transformer (ViViT) \cite{arnab-iccv2021-vivit}, which focuses on capturing the spatiotemporal features of video. However, these model are vulnerable to adversarial examples \cite{goodefellow-iclr2015-fgsm}, \ie, adding human imperceptible perturbations to clean samples, which fool the classification model to make incorrect predictions. This raises the important security concerns from both academia and industry.  

\begin{figure}[!t]
	\centering
	\includegraphics[width=0.90\linewidth]{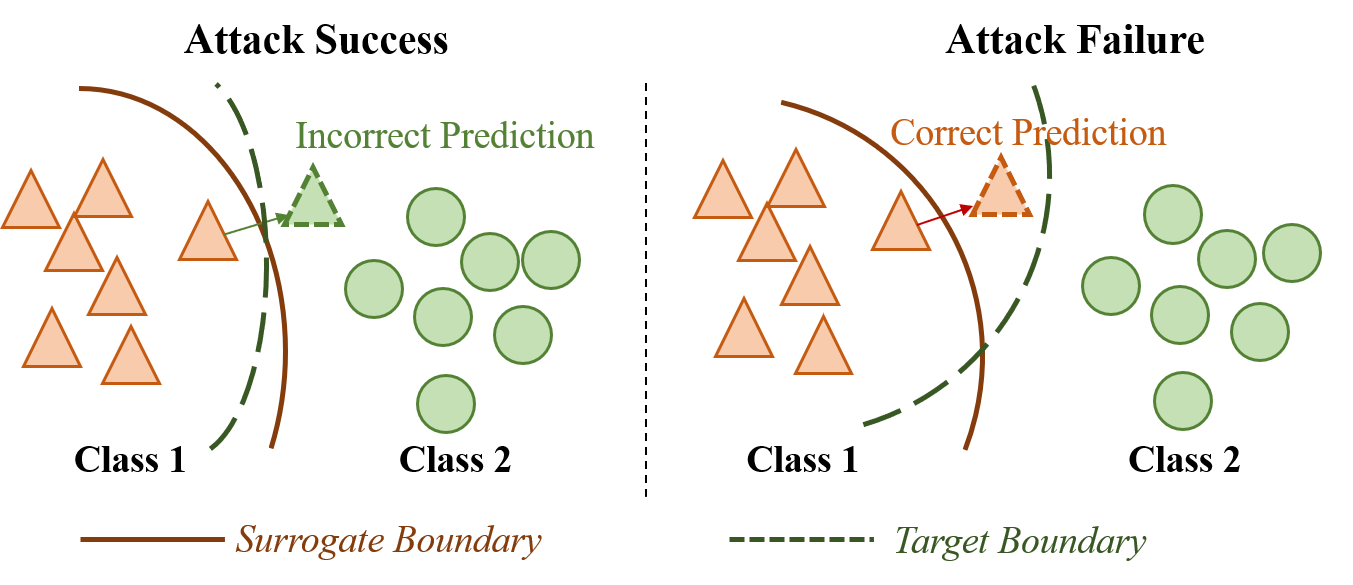}
	\caption{Illustration of the decision boundaries of the surrogate model and the target model. }
	\label{fig:motivation1}
	\vspace{-4mm}
\end{figure}

\begin{figure}[!t]
	\centering
	\includegraphics[width=0.65\linewidth]{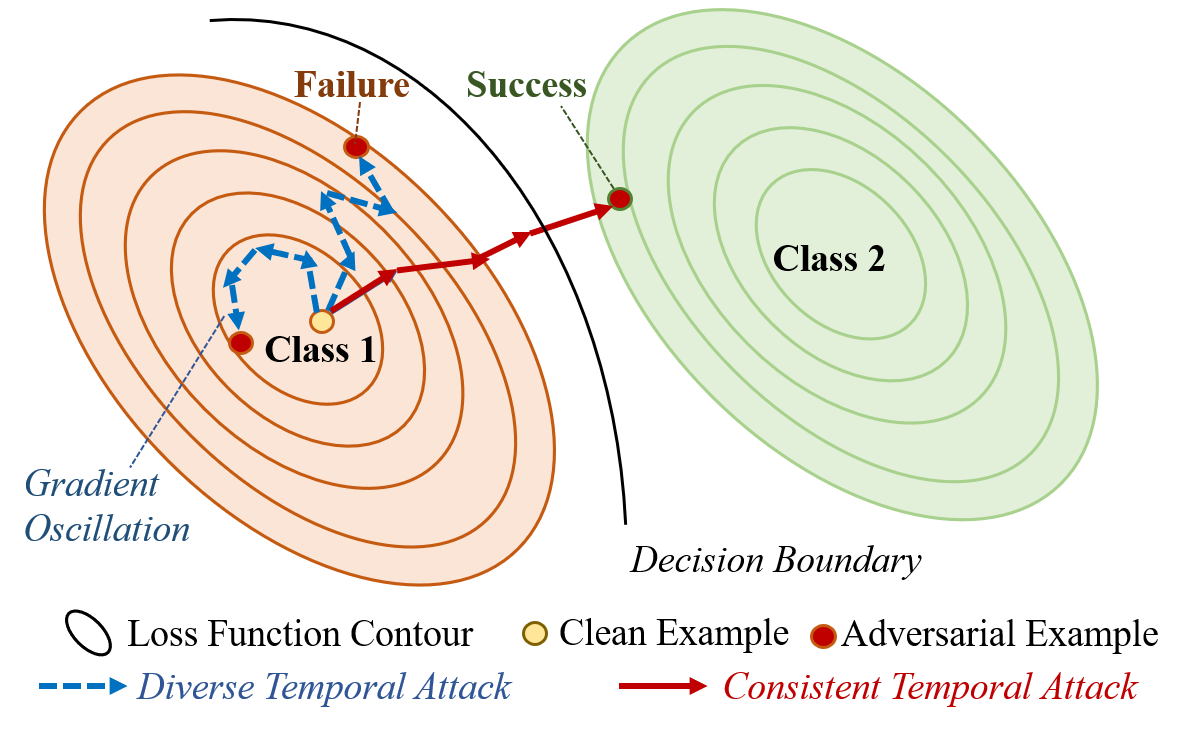}
	\caption{Illustration of diverse and consistent temporal attack. }
	\label{fig:motivation2}
	\vspace{-4mm}
\end{figure}

In real-world scenarios, due to some privacy issue, the prior knowledge such as the parameters and the architecture of target model (\ie, attacked model) are often unavailable to the attacker. Hence, we investigate the transferability of adversarial examples by adopting one white-box model as surrogate or source model to realize black-box attacks for action recognition models. To this end, there are only a few studies devoted to the transfer-based attack on video recognition models. For example, \cite{wei-aaai2022-tt} presents a Temporal Translation (TT) attack method that optimizes the adversarial perturbations on the temporal translated video clips to make the attack less sensitive to varying temporal patterns. From the cross-modal transfer perspective of image-to-video, \cite{wang-aaai2023-i2v} improves the transferability of adversarial examples in black-box scenario by introducing global inter-frame interaction into attack process and disrupting the inherently local correlations of frames within a video; \cite{wei-tpami2024-i2v} generates adversarial examples from white-box image models to attack video models, and optimize perturbations via reducing the similarity of intermedia features between clean frame and adversarial frame. However, the above methods have \textit{two major drawbacks} including: 1) they update gradients to generate adversarial examples which makes them heavily rely on the assumption that the decision boundaries of the surrogate (\aka, source) model and the target model are similar, \ie, the transfer-based attack will fail (\eg, adversarial example is correctly recognized by target model) when the two boundaries are isolated away (see Fig.~\ref{fig:motivation1}), and 2) the existence of the boundary difference incurs the uncertainty problem of attack direction, and they make the attempt in many possible directions, resulting in possible large gradient oscillation as shown in Fig.~\ref{fig:motivation2} and thus weakening the adversarial attack. 

For the \textit{first} problem, from the input transformation perspective, we adopt the adversarial mixup strategy to generate adversarial examples, which makes the transferable attack less dependable on the assumption of the decision boundary similarity between the surrogate model and the target model. The so-called \textit{mixup} acts as data augmentation that linearly interpolates two images and corresponding labels. Here we do not change the label as in \cite{wang-iccv2021-admix}. By contrast, we consider the semantic perturbation on clean sample by making a set of background frames, each of which corresponds to one category. In particular, we add the background frame with the highest attack reward to the clean frame. All clean frames within a video may be mixed up with different background frames. To identify the background frame with the strongest attack ability, we leverage reinforcement learning by designing a reward function for the adversarial Mixer, which includes attack success award, transfer award, and temporal background consistency award. The background frame with the top score is selected for mixing up with the clean frame. Here the temporal background consistency award makes the selected background frames be similar or the same for nearby frames. These skills are wrapped in the model-agnostic \textit{Background Adversarial Mixup} (BAM) module.

For the \textit{second} problem, from the temporal attack perspective, some work \cite{wei-aaai2022-tt} presents a gradient-based temporal translation attack that optimizes the adversarial perturbations on temporally translated video clips, while some others \cite{wei-tpami2024-i2v}\cite{chen-acmmm2023-gcma} focus on feature-based attacks that minimize the cosine similarities between the intermediate features of the (warped) clean frame and the adversarial counterparts. However, they fail to consider the local relation of nearby frames. Meanwhile, \cite{wang-aaai2023-i2v} disrupts the temporal local correlations by reducing the similarity of the adversarial examples of nearby frames, but it introduces the large diversity of adversarial frames by minimizing their feature cosine similarity. This makes the attack directions vary greatly, leading to possible inverse gradient directions as shown in Fig.~\ref{fig:motivation2}. Thus, it adds the difficulty in pushing the adversarial example away from the current decision boundary. Hence, we design the \textit{Background-induced Temporal Gradient enhancement} (BTG) module that leverages the background attack loss and the temporal gradient consistency loss to make nearby frames have similar gradients. This makes the attack directions be consistent across the frames along the temporal dimension, enhancing the attack ability of adversarial example gradually. 


We briefly summarize the main contributions as follows:
 
\begin{itemize}
	\item We study the transfer-based black-box attack on action recognition models, and propose a \textbf{B}ackground \textbf{M}ixup-induced \textbf{T}emporal \textbf{C}onsistency (\textbf{BMTC}) attack method to boost the transferability of adversarial examples.
	
	\item We perform a model-agnostic input transformation by considering background semantics, \ie, adversarially mixup background frame from other categories with clean frame, while that background frame is determined by the reward function using reinforcement learning with good transferability and temporal consistency. 
	
	\item We strengthen the attack direction across sequential frames in a progressive way, by imposing the temporal gradient consistency constraint on the loss and guiding the attack in the direction of the background category. 
	
\end{itemize}

\section{Related Works}

\subsection{Transfer-based Image Attacks}
Previous methods often adopt white-box attacks, such as Fast Gradient Sign Method (FGSM) \cite{goodefellow-iclr2015-fgsm} that updates the gradient on the clean sample to maximize the loss function, Projected Gradient Descent (PGD) \cite{madry-iclr2018-pgd} that is an iterative FGSM. To improve the transferability of adversarial examples on black-box models, there are three strategies: 1) \textit{data augmentation} (\ie,input transformation), \eg, diversity input attack \cite{xie-cvpr2019-dim}, Scale-Invariant Method (SIM) \cite{lin-iclr2020-sini}, Translation-Invariant (TI) attack \cite{dong-cvpr2019-tim}, Adversarial Mixup (AdMix) \cite{wang-iccv2021-admix} that mixes two images using Mixup \cite{zhang-iclr2018-mixup}, and Path-Augmented Method (PAM) \cite{wang-cvpr2023-pam} that constructs a candidate augmentation path pool for path selection with greedy search; 2) \textit{gradient modification}, \eg, Momentum Iterative (IM) attack \cite{dong-cvpr2018-mifgsm}, Skip Gradient Method (SGM) \cite{wu-iclr2020-sgm}, Variance Momentum Iterative tuning (VMI) \cite{wang-cvpr2021-vmi}, and Gradient-Related Adversarial Attack (GRA) \cite{zhu-iccv2023-gra}; 3) \textit{feature disruption}, \eg, Intermediate Level Attack (ILA) \cite{huang-iccv2019-ila}, Feature Importance-Aware attack (FIA) \cite{wang-iccv2021-fia}, and ILA with Data Augmentation (ILA-DA) \cite{yan-iclr2023-ilada}. 


\subsection{Transfer-based Video Attacks}
Transferable attacks on video models are less explored compared to that on image models. Existing methods attempt to break the temporal relations, \eg, from the gradient update perspective, Temporal Translation (TT) \cite{wei-aaai2022-tt} does multiple translations on frames along the temporal dimension to generate the clips under different translation patterns and optimizes the perturbations on these clips; from the feature disruption perspective, some methods minimize the cosine similarities between the intermediate features of the clean frames \cite{wei-tpami2024-i2v} or the warped clean frames \cite{chen-acmmm2023-gcma} and the adversarial counterparts. Nevertheless, they treat video as a collection of unordered images, ignoring the inherent temporal continuity. In contrast, \cite{wang-aaai2023-i2v} disrupts the temporal local correlations of nearby frames by reducing the similarity of the adversarial counterparts, and minimizes the cosine similarities among the intermediate features of the benign points lying on a convex hull, which leads to diverse adversarial frames. This may incur gradient oscillation, \eg, some sample gradients update in reverse direction, because the uncertain attack directions vary greatly. 

\subsection{Action Recognition Models}
Much progress has been made in deep neural network based action recognition models. Existing methods have two groups: 1) Convolutional Neural Networks (CNNs) methods, \eg, Inflated 3D ConvNet (I3D) \cite{carreira-cvpr2017-kinetics}, Non-Local neural networks (NL) \cite{wang-cvpr2018-nl}, SlowFast \cite{feichtenhofer-iccv2019-slowfast} that captures spatial semantics at low frame rate and the motion dynamics at fine temporal resolution, Temporal Pyramid Networks (TPN) \cite{yang-cvpr2020-tpn} that uses the source features and the fusion features to form a feature hierarchy to capture action instances at various tempos; 2) Vision Transformer (ViT) methods, \eg, VideoTransformer
Network (VTN) \cite{neimark-cvpr2021-videotransformer} that is built on top of any given 2D spatial network and attends to the entire video, Time-Space Transformer (TimeSformer) \cite{bertasius-icml2021-video} that leverages 3D self-attention over the space-time volume to capture long-range temporal dependency among frames, Motionformer \cite{patrick-nips2021-motionformer} aggregates implicitly motion path information to model the temporal dynamic scenes, and VideoSwin \cite{liu-cvpr2022-videoswin} that encourages an inductive bias of locality by adapting the Swin Transformer \cite{liu-iccv2021-swin} for a better speed-accuracy trade-off.
\\

\begin{figure*}[!t]
	\centering
	\includegraphics[width=0.9\linewidth]{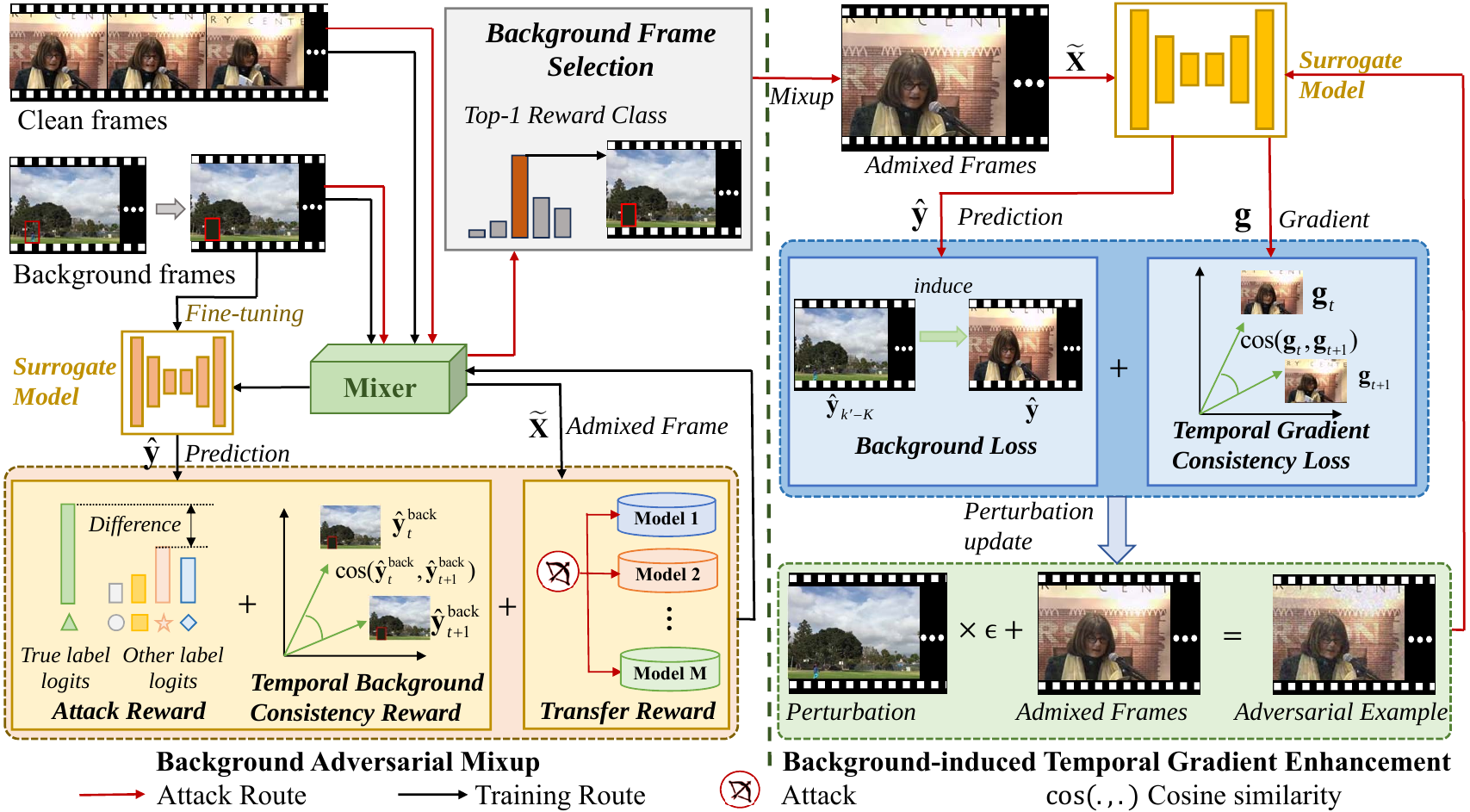}
	\caption{Overall framework of our Background Mixup-induced Temporal Consistency (BMTC) attack method for action recognition.}
	\label{fig:framework}
	\vspace{-4mm}
\end{figure*}

\section{Methodology}
This section introduces the proposed Background Mixup-induced Temporal Consistency (BMTC) attack method in the transfer-based black-box setting for action recognition models. The overall framework is illustrated in Fig.~\ref{fig:framework}, which consists of two primary components, \ie, Background Adversarial Mixup (BAM) module and Background-induced Temporal Gradient enhancement (BTG) module.

\subsection{Problem Definition}
Given a video sample $x$ represented by a tensor $\mathbf{X} \in \mathbb{R}^{T\times H\times W\times C}$ with the ground-truth label $y \in \mathcal{Y}=\{1, 2, \cdots, K\}$ represented by a one-hot vector $\mathbf{y}\in \mathbb{R}^K$, where $\{N, H, W, C\}$ denote the number, height, width, and channel of the video, each frame is indexed by $t\in \{1, 2, \cdots, T\}$, and there are $K$ action categories, transfer-based adversarial attack aims to generate an adversarial example $\mathbf{X}^{adv}=\mathbf{X}+\delta$ by the surrogate model $g(\cdot)$ to fool the target model $f(\cdot): \mathbf{X} \mapsto \mathcal{Y}$ to make incorrect predictions, \ie, $f(\mathbf{X}^{adv}) \neq y$ without knowing the gradients and architecture of the model, where the perturbation $\delta$ is restricted by the $\ell_p$-norm $\|\delta\|_p \le \epsilon$. Here, $\epsilon>0$ is a constant that governs the perturbation magnitude, and we adopt $\ell_{\inf}$-norm and untargeted adversarial attacks as in \cite{wei-tpami2024-i2v}\cite{wei-aaai2022-tt}. The objective of untargeted adversarial attack is formulated as:
\begin{equation}
	\label{eq:obj_untarget_attack_white}
	\arg \max_{\delta} J(f(x +\delta), y), \textit{s.~t.~} \|\delta\|_{\inf} \le \epsilon,
\end{equation}
where the function $J(\cdot)$ often adopts cross-entropy loss. Since this work focuses on the black-box transfer-based attack, the attacker (\aka, adversary) has no access to details of the target model $f(\cdot)$. We aim to improve the black-box transferability of video adversarial examples on other action recognition models.  

\subsection{Background Adversarial Mixup}
To ensure the good transferability of attacks, the decision boundary of the target model is expected to approach that of the surrogate model. When their decision boundaries are far away, it results in poor transferability of adversarial examples, \eg, they make different predictions on the same sample. Gradient modification methods heavily depends the above decision boundary assumption, since the gradients back-propagated during training have essential compacts on reshaping the decision boundary of the action recognition model. To circurvement this problem, we present a model-agnostic input transformation method, which adopts the adversarial mixup with background in video using a Mixer that consists of a feature extractor (\eg, ResNet50) and a classification head (\eg, fully-connected layer). The rationale behind this is that there exist strong correlations between action category and action background, \eg, \textit{surfing} in the blue sea and \textit{riding} on a road, which makes it possible that mixing the background from other categories with a given video might mislead the model to yield wrong predictions. This raises a key problem that how to select the background frame. The working mechanism is shown below.

We construct a set of candidate background frames, which are obtained by randomly selecting one background frame of a randomly chosen video from each category. The selected background frame is made by applying one zero-shot video object segmentation method, \ie, Isomerous transformer \footnote{https://github.com/DLUT-yyc/Isomer} \cite{yuan-iccv2023-isomervos}, to abandon foreground part, such as person and action-related objects. Note that the label of each background frame is kept the same as that in the original video, and there is a one-to-one correspondence of label mapping between original video and background frame, \ie, $y^{back} = y +K $, whose one-hot vector is $\mathbf{y}^{back}\in \mathbb{R}^K$ and its index is $k^\prime = \{1, 2, \cdots, K\}$. The background frame is repeated to form the background video $\mathbf{X}^{back} \in \mathbb{R}^{T\times H\times W\times C}$ for a video $\mathbf{X}$. These background videos are added to the set of original videos and they are fed into the surrogate model for fine-tuning, whose loss $\mathcal{L}_{ft}$ is defined as
\begin{equation}
	\label{eq:obj_finetune}
	\mathcal{L}_{ft} = -\sum_{k=1}^K \mathbf{y}_k \log (\hat{\mathbf{y}}_k) - \lambda \sum_{k^\prime = 1}^K \mathbf{y}^{back}_{k^\prime} \log (\hat{\mathbf{y}}^{back}_{k^\prime}),
\end{equation}
where the first term is action loss and the second term is background loss, $\{\hat{\mathbf{y}}_k, \hat{\mathbf{y}}^{back}_{k^\prime}\}$ are the probability logits from the model, the hyper-parameter $\lambda$ (empirically set to 0.2) is used to balance the contribution of the two terms to the objective. Note that fine-tuning the surrogate model is to equip the surrogate model with some discriminative ability of different background categories, so the action loss should play a dominant role and the background loss serves as an auxiliary.

To select the background frame with the most powerful attack ability, we adopt the reinforcement learning to compute three reward functions, \ie, attack reward $R_{attack}$, transfer reward $R_{transfer}$, and temporal background consistency reward $R_{tbc}$. These reward functions are used to encourage the Mixer to generate the admixed sample to successfully attack the surrogate model. Here the Mixer is used to mixup the selected background video with the original video, \ie, $\tilde{\mathbf{X}} = (1-\gamma)\cdot \mathbf{X} + \gamma\cdot \mathbf{X}^{back}$, where the constant $\gamma$ is empirically set to 0.2. For the frame-level mixup, the $i$-th admixed frame is $\tilde{\mathbf{X}}_i = (1-\gamma)\cdot \mathbf{X}_i + \gamma\cdot \mathbf{X}^{back}_i$. Actually, admixed sample serves as the vanilla adversarial example. 

\textbf{Attack Reward}. For a video $x$ with its label $y$ and the predicted probability vector $\hat{\mathbf{y}}=[\hat{y}_1, \hat{y}_2, \cdots, \hat{y}_K]$, this reward $R_{attack}$ takes the difference of the label prediction probabilities before and after the adversarial mixup, \ie, 
\begin{equation}
	\label{eq:reward_attack}
	R_{attack} = \max \{\hat{y}_{\backslash k}\} - \hat{y}_k,~ \textit{s.t.}~k=y,
\end{equation}
where $\{\hat{y}_{\backslash k}\}$ is a set excluding the $k$-th entry, which indicates the true label $k$. When the probability of the true label $\hat{y}_k$ increases, the reward score decreases; when the maximum probability of other categories $\max \{\hat{y}_{\backslash k}\}$ increases, the reward score rises, \ie, when the video is incorrectly predicted, we should give the model positive reward.

\textbf{Transfer Reward}. To improve the transferability of adversarial example across different models, we provide the transfer reward $R_{transfer}$ to attack $M$ black-box target models, and averages their attack rewards, \ie,
\begin{equation}
	\label{eq:reward_transfer}
	R_{transfer} = \frac{1}{K} \sum_{m=1}^M R^m_{attack},
\end{equation}
where $R^m_{attack}$ is the attack award of the $m$-th target model.

\textbf{Temporal Background Consistency Reward}. Since similar samples have similar backgrounds, nearby frames are expected to have similar backgrounds. For a video $x$ with $T$ frames indexed by $t$, the neighbouring frames are expected to be mixed with similar or the same background frames, \ie, temporal consistency among background frames, such that the attack strength will be boosted. We define the Temporal Background Consistency (TBC) reward $R_{tbc}$ as
\begin{equation}
	\label{eq:reward_temporal}
	R_{tbc} = \frac{1}{T - 1} \sum_{t = 1}^{T - 1} \frac{\hat{\mathbf{y}}^{back}_t\cdot \hat{\mathbf{y}}^{back}_{t+1}}
		{\|\hat{\mathbf{y}}^{back}_t\|_2\cdot \|\hat{\mathbf{y}}^{back}_{t+1}\|_2},
\end{equation}
where $\|\cdot\|_2$ denotes the $\ell_2$-norm, and the vector $\hat{\mathbf{y}}^{back}_t \in \mathbb{R}^K$ is the predicted probability of the $t$-th background frame. 

Therefore, the total reward of selecting the background frame is 
\begin{equation}
	\label{eq:reward_total}
	R_{total} =  R_{attack} + \alpha_1 \times R_{transfer} + \alpha_2 \times R_{tbc}
\end{equation}
where the constants $\alpha_1>0$ (set to 0.3) and $\alpha_2>0$ (set to 0.1) are used to control the contributions of the transfer reward and the temporal background consistency reward, respectively. Finally, we select the background frame with the highest total reward score from $K$ candidates to mixup with the original video frames sequentially. 

\subsection{Background-induced Temporal Gradient Enhancement}
To further improve the transferability of vanilla adversarial example (\ie, admixed sample), we introduces two adversarial losses including the background attack loss and the temporal gradient consistency loss. The former is used to guide the surrogate model to generate the adversarial example, by learning the probability logits along the direction of the original action category, which is associated with the mixed background frame. In particular, we maximize the cross-entropy loss of other categories and minimize that of the background category (\ie, fool the model to wrongly predict the sample as the selected background category), \ie,
\begin{equation}
	\label{eq:loss_background}
	\mathcal{L}_{back} = -\sum_{j=1, j\ne k^\prime\text{-}K}^K \mathbf{y}_j \log \hat{\mathbf{y}}_j + \mathbf{y}_{k^\prime\text{-}K} \log \hat{\mathbf{y}}_{k^\prime\text{-}K},
\end{equation}
where the background frame belongs to the $(k^\prime-K)$-th category from the original $K$ actions. 

To makes the attack direction be consistent between nearby adversarial frames, we design the Temporal Gradient Consistency (TGC) loss to encourage the surrogate model to generate the adversarial counterparts with similar gradients for nearby frames. By this means, the attack strength is gradually enhanced across the frames along the temporal dimension. Mathematically, we adopt the cosine similarity to evaluate the temporal consistency between the gradients of two nearby adversarial frames, \ie,
\begin{equation}
	\label{eq:loss_tgc}
	\mathcal{L}_{tgc} = \frac{1}{T- 1} \sum_{t=1}^{T-1} \frac{\mathbf{g}^{t^\top} \cdot \mathbf{g}^{t+1}}
	     {\| \mathbf{g}^t \|_2 \cdot  \| \mathbf{g}^{t+1} \|_2},
\end{equation}
where $\mathbf{g}^t \in \mathbb{R}^{H\cdot W \cdot 3}$ denotes the gradient vector of the $t$-th adversarial frame $\mathbf{X}_t^{adv}\in \mathbb{R}^{H\times W\times C}$, which is calculated as
\begin{equation}
	\label{eq:gradient}
	g^t = \nabla_{\mathbf{X}^{adv}_t} \mathcal{L}_{back}(\mathbf{X}^{adv}_t, \mathbf{y}; \mathbf{\theta}), \mathbf{g}^t = \text{vec}(g^t),
\end{equation}
where $\theta$ denotes the model parameter set, $\text{vec}(\cdot)$ vectorizes a tensor to the corresponding vector, and $\nabla_{\mathbf{X}^{adv}_t} L$ is the gradient of the loss $\mathcal{L}_{back}$ \wrt the $t$-th frame.

Therefore, the total temporal adversarial attack loss is: 
\begin{equation}
	\label{eq:loss_total}
	\mathcal{L}_{total} = \mathcal{L}_{back} + \beta\mathcal{L}_{tgc},
\end{equation}
where the hyper-parameter $\beta\in [0,1]$ (set to 0.1) controls the balance between the two terms during optimization. 

\subsection{Adversarial Example Generation}
This work adopts the Projected Gradient Descent (PGD) algorithm \cite{madry-iclr2018-pgd} to compute the sample gradients of the model, and the adversarial example at the $i$-th iteration is updated in the following way:
\begin{equation}
	\label{eq:adversary}
	\mathbf{X}^{i+1}=\Pi_{\mathbf{X},\epsilon}\left[\mathbf{X}^i+\eta\cdot \text{sign}\left(\nabla_{\mathbf{X}^i} \mathcal{L}_{total}(\mathbf{X}^i, \mathbf{y})\right)\right],
\end{equation}
where $\eta>0$ is the step size, $\Pi_{\mathbf{X}, \epsilon}[\cdot] = \min(\max(\mathbf{X}^i, \mathbf{X}-\epsilon), \mathbf{X}+\epsilon)$ is the projection function that controls the perturbation magnitude, $\mathbf{X}$ is the clean sample, and the initial $\mathbf{X}^0$ is the admixed sample. When the iteration achieves the maximum number $N_{iter}$ (\eg, set to 10 for a good tradeoff between performance and efficiency), we obtain the ultimate adversarial example to attack the target models. %

\section{Experiments}
All experiments were performed on a server equipped with two 48G NVIDIA A6000 graphics cards. The codes are compiled with PyTorch 1.11, Python 3.8, and CUDA 11.4. Our code is available at {\b \href{https://github.com/mlvccn/BMTC\_TransferAttackVid}{Github}}.

\subsection{Datasets and Evaluation Metric}
We conduct experiments on two video benchmarks including UCF101 \cite{soomro-arxiv2012-ucf101} and Kinetics-400  \cite{carreira-cvpr2017-kinetics}, and one image benchmark ImageNet \cite{deng-cvpr2009-imagenet}. Following \cite{wei-aaai2022-tt}\cite{wei-tpami2024-i2v}, we adopt the Attack Success Rate (ASR) as the metric, which calculates the rate of adversarial examples misclassified by the black-box model. The higher the ASR, the better the transferability is. As a common practice, one video is randomly selected from each category and correctly classified by all target models. 


\subsection{Experimental Settings}
For the adversarial Mixer, we train a classification model from scratch with ResNet50 as backbone on Kinetics-400 and ImageNet respectively, and the training epochs are 100. For the surrogate model fine-tuning, we adopt TPN and ResNet101 as the surrogate of video (20 epochs) and image (10 epochs) attack, respectively, and the hyper-parameter of the fine-tuning loss is $\lambda=0.2$. For both of them, the initial learning rate is 0.1, momentum is 0.9, and the weight decay is 1e4. For the adversarial example generation, we set the maximum perturbation $\epsilon$ to 16 for video and 8 for image, the attack step size $\eta$ to 1.6 for video and 0.8 for image, the maximum iteration number $N_{iter}$ to 10.
 
For action recognition, we examine three CNN models with different architectures, \ie, Non-local network (NL) \cite{wang-cvpr2018-nl}, SlowFast (SF) \cite{feichtenhofer-iccv2019-slowfast}, and Temporal Pyramid Networks (TPN) \cite{yang-cvpr2020-tpn}, which employ ResNet50 and ResNet101 as backbones. We train video models from scratch with Kinetics-400 and fine-tune them on UCF101. Following \cite{wei-tpami2024-i2v}, we skip every other frame from randomly selected clip with 64 consecutive frames to make input for Kinetics-400, and use 32 consecutive frames as input clips for UCF101. Moreover, we also examine four Transformer models including VideoTransformer Network (VTN) \cite{neimark-cvpr2021-videotransformer}, Time-Space Transformer (TimeSformer) \cite{bertasius-icml2021-video}, Motionformer \cite{patrick-nips2021-motionformer}, and VideoSwin \cite{liu-cvpr2022-videoswin}. Following \cite{wei-tpami2024-i2v}, we sample one clip per video and each clip consists of 16 frames with the temporal stride 4. For image classification, we examine four models including ResNet18 \cite{he-cvpr2016-resnet}, ResNet101, ResNeXt50 (RNX50) \cite{xie-cvpr2017-resnext}, and DenseNet121 (DN121) \cite{huang-cvpr2017-densenet}. Note that NL-101, SF-101 and TPN-101 using ResNet101 act as white-box models (surrogate) for generating adversarial examples to attack other black-box models (target). The input spatial size is 224$\times$224. The model number $M$ is 3 for video and 4 for image.

\begin{table}[!t]
	\centering
	\caption{Performance comparison on UCF101.}
	\vspace{-2mm}
	\label{tbl:ucf101}
	\scalebox{0.7}{   
		\setlength{\tabcolsep}{0.8mm} 
		\begin{tabular}{lllcccccccccccccc}
			\toprule[0.75pt]
			\multirow{2}{*}{Surrogate} & \multirow{2}{*}{Attack} & \multirow{2}{*}{Venue} & \multicolumn{4}{c}{NL}        & \multicolumn{4}{c}{SlowFast}        & \multicolumn{4}{c}{TPN}           \\ 
			\cmidrule(lr){4-7} \cmidrule(lr){8-11} \cmidrule(lr){12-15}
			&                            &                        & & RN101 & RN50 & & & RN101 & RN50 & & & RN101 & RN50 & \\ 
			\midrule[0.5pt]
			\multirow{4}{*}{NL-101}     & PGD    & ICCV'18      & & \cellcolor{gray!20}92.08 & \cellcolor{gray!20}31.68 & & & 11.88 & 15.84 & & & 8.91 & 10.89 &  \\ 
			& AA     & ICML'20     & & \cellcolor{gray!20}94.05 & \cellcolor{gray!20}28.71 & & & 17.82 & 20.79 & & & 12.87 & 11.88 &  \\ 
			& TT     & AAAI'22     & & \cellcolor{gray!20}81.19 & \cellcolor{gray!20}62.38 & & & 48.52 & 58.42 & & & 37.62 & 39.60 &  \\ 
			& Ours   & -           & & \cellcolor{gray!20}88.12 & \cellcolor{gray!20}79.21 & & & \textbf{71.28} & \textbf{69.31} & & & \textbf{53.47} & \underline{54.45} &  \\ 
			\midrule[0.5pt]
			\multirow{4}{*}{SF-101} & PGD    & ICCV'18     & & 17.82 & 23.76 & & & \cellcolor{gray!20}93.07 & \cellcolor{gray!20}36.63 & & & 9.90 & 14.85 &  \\ 
			& AA     & ICML'20     & & 19.80 & 21.78 & & & \cellcolor{gray!20}93.07 & \cellcolor{gray!20}35.64 & & & 16.83 & 18.81 &  \\ 
			& TT     & AAAI'22     & & 54.46 & 62.38 & & & \cellcolor{gray!20}80.10 & \cellcolor{gray!20}52.48 & & & 36.63 & 38.61 &  \\ 
			& Ours   & -           & & \textbf{73.26} & \textbf{78.22} & & & \cellcolor{gray!20}86.14 & \cellcolor{gray!20}71.29 & & & \underline{52.47} & \textbf{66.34} &  \\ 
			\midrule[0.5pt]
			\multirow{4}{*}{TPN-101}    & PGD    & ICCV'18     & & 11.88 & 10.89 & & & 9.90 & 14.85 & & & \cellcolor{gray!20}84.16 & \cellcolor{gray!20}33.67 &  \\ 
			& AA     & ICML'20     & & 13.86 & 17.82 & & & 10.89 & 18.81 & & & \cellcolor{gray!20}89.11 & \cellcolor{gray!20}24.75 &  \\ 
			& TT     & AAAI'22     & & 59.41 & 62.37 & & & 43.56 & 44.55 & & & \cellcolor{gray!20}78.22 & \cellcolor{gray!20}35.64 &  \\ 
			& Ours   & -           & & \underline{69.31} & \underline{76.24} & & & \underline{64.35} & \underline{67.33} & & & \cellcolor{gray!20}82.17 & \cellcolor{gray!20}73.27 &  \\ 
			\toprule[0.75pt]
		\end{tabular}
	}
\end{table}

\begin{table}[!t]
	\centering
	\caption{Performance comparison on Kinetics-400.}
		\vspace{-2mm}
	\label{tbl:kinetics}
	\scalebox{0.7}{   
		\setlength{\tabcolsep}{0.8mm}{ 
			\begin{tabular}{lllcccccccccccccc}
				\toprule[0.75pt]
				\multirow{2}{*}{Surrogate} & \multirow{2}{*}{Attack} & \multirow{2}{*}{Venue} & \multicolumn{4}{c}{NL}        & \multicolumn{4}{c}{SlowFast}        & \multicolumn{4}{c}{TPN}           \\ 
				\cmidrule(lr){4-7} \cmidrule(lr){8-11} \cmidrule(lr){12-15}
				&                            &                        & & RN101 & RN50 & & & RN101 & RN50 & & & RN101 & RN50 & \\ 
				\midrule[0.5pt]
				\multirow{4}{*}{NL-101}     & PGD    & ICCV'18      & & \cellcolor{gray!20}94.75 & \cellcolor{gray!20}13.75 & & & 13.50 & 16.25 & & & 11.25 & 14.25  \\ 
				& AA     & ICML'20     & & \cellcolor{gray!20}99.00 & \cellcolor{gray!20}25.25 & & & 20.25 & 21.25 & & & 14.25 & 20.75  \\ 
				& TT     & AAAI'22     & & \cellcolor{gray!20}97.25 & \cellcolor{gray!20}77.25 & & & 78.25 & 75.75 & & & 57.25 & 62.00  \\ 
				& Ours   & -           & & \cellcolor{gray!20}98.50 & \cellcolor{gray!20}85.25 & & & \underline{80.50} & \textbf{83.00} & & & \textbf{75.25} & \underline{78.50}  \\ 
				\midrule[0.5pt]
				\multirow{4}{*}{SF-101} & PGD    & ICCV'18     & & 20.25 & 25.75 & & & \cellcolor{gray!20}91.00 & \cellcolor{gray!20}36.25 & & & 15.75 & 14.85  \\ 
				& AA     & ICML'20     & & 26.50 & 30.00 & & & \cellcolor{gray!20}93.25 & \cellcolor{gray!20}40.75 & & & 19.50 & 25.50  \\ 
				& TT     & AAAI'22     & & 59.75 & 61.25 & & & \cellcolor{gray!20}94.00 & \cellcolor{gray!20}75.25 & & & 53.50 & 62.75  \\ 
				& Ours   & -           & & \textbf{67.00} & \underline{70.25} & & & \cellcolor{gray!20}94.50 & \cellcolor{gray!20}83.50 & & & \underline{74.25} & \textbf{83.75} \\ 
				\midrule[0.5pt]
				\multirow{4}{*}{TPN-101}    & PGD    & ICCV'18     & & 14.00 & 10.25 & & & 15.50 & 14.85 & & & \cellcolor{gray!20}94.00 & \cellcolor{gray!20}33.50  \\ 
				& AA     & ICML'20     & & 19.75 & 21.50 & & & 24.75 & 31.50 & & & \cellcolor{gray!20}99.00 & \cellcolor{gray!20}24.75  \\ 
				& TT     & AAAI'22     & & 49.75 & 57.50 & & & 69.25 & 66.00 & & & \cellcolor{gray!20}95.50 & \cellcolor{gray!20}89.25  \\ 
				& Ours   & -           & & \underline{63.25} & \textbf{70.75} & & & \textbf{81.50} & \underline{80.25} & & & \cellcolor{gray!20}98.50 & \cellcolor{gray!20}84.25  \\ 
				\midrule[0.5pt]
			\end{tabular}
		}
	}
\end{table}

\begin{table}[!t]
	\centering
	\caption{Performance comparison on Kinetics-400 (ViT models).}
		\vspace{-2mm}
	\label{tbl:kinetics_vit}
	\scalebox{0.75}{   
		\setlength{\tabcolsep}{1.0mm}{
			\begin{tabular}{llcccc}
				\toprule[0.75pt]
				Attack & Surrogate & VTN & TimeSformer & Motionformer & VideoSwin \\ 
				\midrule[0.5pt]
				\multirow{3}{*}{TT} & NL & 46.75 & 40.75 & 36.25 & 41.50 \\
				& SlowFast & 49.50 & 45.50 & 41.50 & 32.25 \\
				& TPN & 51.25 & 37.75 & 44.00 & 41.50 \\ 
				\midrule[0.5pt]
				ENS-I2V-MF & Ensemble & 53.50 & 42.00 & 36.75 & 56.25 \\ 
				\midrule[0.5pt]
				AENS-I2V-MF & Ensemble & 54.00 & 43.75 & 39.50 & 55.50 \\ 
				\midrule[0.5pt]
				\multirow{3}{*}{Ours} & NL & 67.25 & 61.25 & \underline{59.25} & \textbf{69.75} \\
				& SlowFast & \underline{69.00} & \underline{65.75} & 57.75 & 66.50 \\
				& TPN & \textbf{72.50} & \textbf{66.50} & \textbf{60.00} & \underline{68.25} \\ 
				\toprule[0.75pt]
			\end{tabular}
		}
	}
\end{table}

\subsection{Quantitative Results}
\textbf{Action Recognition}. The attack results of action recognition models are reported in Table~\ref{tbl:ucf101} on UCF101, Tables~\ref{tbl:kinetics} and \ref{tbl:kinetics_vit} on Kinetics-400. The best records are highlighted in boldface, and the second best ones are underlined. Here, ``RN'' denotes ResNet, the records with the gray color are the results of white-box attack to be overlooked. 

From these tables, we observe that our method achieves more satisfying performance across several action recognition models including NL, SlowFast, and TPN, with different architectures. For example, when using the adversarial example from SF-101 to attack NL and TPN with RN50 as backbone, our method has a gain of 9.0\% and 21.0\% respectively, in comparison to TT on Kinetics-400; when using the adversarial example from TPN-101 to attack NL and SlowFast with RN50 as backbone, our method has an improvement of 13.87\% and 22.78\% respectively, compared to the most competitive alternative TT on UCF101. This demonstrates the superiority of the adversarial mixup strategy with reinforcement learning and the temporal consistency among nearby frames. Besides CNN models, the performance improvements are also found on ViT models in Table~\ref{tbl:kinetics_vit}. 

Meanwhile, the cross-modal attack results in terms of ASR are shown in Table~\ref{tbl:average_res_ucf101} on UCF101 and Table~\ref{tbl:average_res_kinetics} on Kinetics-400. Here, the surrogate models of the compared methods are AlexNet \cite{alex-nips2012-alexnet}, ResNet-101 \cite{he-cvpr2016-resnet}, SqueezeNet \cite{iandola-arxiv2016-squeezenet}, and VGG-16 \cite{simonyan-iclr2015-vgg} in image domain, while we use the other two video models as the surrogate models, \eg, SF and TPN as surrogate and NL as target (group~1). The results are averaged over those surrogate models, which indicates the advantages of the proposed attack method.

\noindent\textbf{Image Classification}. The results of image classification models are reported in Table~\ref{tbl:res_imagenet} on ImageNet. Here, ``Ori'' denotes the vanilla attack method (column~2). Note that the temporal consistency reward and background-induced loss fail in image domain. Even though only using the adverse award and the transfer award to select the background, ours still improves the attack performance of the vanilla ones. This once again validates the power of the adversarial mixup with the background frame.

\begin{table}[!t]
	\centering
	\caption{Average results on UCF101.}
		\vspace{-2mm}
	\label{tbl:average_res_ucf101}
	\scalebox{0.7}{   
		\setlength{\tabcolsep}{2mm}{
			\begin{tabular}{llcccccc}
				\toprule[0.75pt]
				\multirow{2}{*}{Attack} & \multirow{2}{*}{Venue} & \multicolumn{2}{c}{SF/TPN$\rightarrow$NL}                      & \multicolumn{2}{c}{NL/TPN$\rightarrow$SF}          & \multicolumn{2}{c}{NL/SF$\rightarrow$TPN}                    \\ \cmidrule(lr){3-4} \cmidrule(lr){5-6} \cmidrule(lr){7-8}
				&                                 & \multicolumn{1}{c}{RN101} & \multicolumn{1}{c}{RN50} & \multicolumn{1}{c}{RN101} & \multicolumn{1}{c}{RN50} & \multicolumn{1}{c}{RN101} & \multicolumn{1}{c}{RN50} \\ 
				\midrule[0.5pt]
				I2V                                     & CVPR'22                         & 54.20                                  & 37.87                                 & 44.55                                  & 33.91                                 & 46.02                                  & 14.25                                 \\
				GCEC                                   & AAAI'23                         & \underline{63.87}                                  & \underline{44.80}                                 & \underline{52.47}                                 & \underline{38.86}                                 & 52.72                                  & 20.75                                 \\
				I2V-MF                                  & TPAMI'24                        & 57.67                                  & 37.87                                 & 48.26                                  & 23.87                                 & 48.26                                  & \underline{62.00}                                 \\
				Ours                                    &                                 & \textbf{77.23}                                  & \textbf{67.82}                                 & \textbf{68.32}                         & \textbf{52.97}                        & \textbf{60.40}                         & \textbf{78.50}\\ 
				\toprule[0.75pt]                                
			\end{tabular}
		}
	}
\end{table}

\begin{table}[!t]
	\centering
	\caption{Average results on Kinetics-400.}
	\label{tbl:average_res_kinetics}
	\scalebox{0.7}{   
		\setlength{\tabcolsep}{2.2mm}{
			\begin{tabular}{llcccccc}
				\toprule[0.75pt]
				\multirow{2}{*}{Attack} & \multirow{2}{*}{Venue} & \multicolumn{2}{c}{SF/TPN$\rightarrow$NL}                      & \multicolumn{2}{c}{NL/TPN$\rightarrow$SF}          & \multicolumn{2}{c}{NL/SF$\rightarrow$TPN}                    \\ \cmidrule(lr){3-4} \cmidrule(lr){5-6} \cmidrule(lr){7-8}
				&                                 & \multicolumn{1}{c}{RN101} & \multicolumn{1}{c}{RN50} & \multicolumn{1}{c}{RN101} & \multicolumn{1}{c}{RN50} & \multicolumn{1}{c}{RN101} & \multicolumn{1}{c}{RN50} \\ 
				\midrule[0.5pt]
				I2V                                     & CVPR'22  & 44.25 & 54.13 & 64.13 & 63.94 & 65.38 & 72.19 \\
				GCEC                                   & AAAI'23    & \underline{54.56} & \underline{63.88} & \underline{71.88} & \underline{70.63} & \underline{72.19} & \underline{77.31} \\
				I2V-MF                                  & TPAMI'24    & 46.44 & 55.31 & 66.06 & 65.63 & 69.75 & 74.63 \\
				Ours                                    &                                 & \textbf{70.50}                         & \textbf{81.00}                        & \textbf{81.63}                         & \textbf{74.75}                        & \textbf{81.13}                         & \textbf{78.50}\\ 
				\toprule[0.75pt]                       
			\end{tabular}
		}
	}
\end{table}

\begin{table}[!t]
	\caption{Performance comparison on ImageNet.}
	\label{tbl:res_imagenet}
	\scalebox{0.6}{   
		\setlength{\tabcolsep}{0.6mm}{
			\begin{tabular}{lllcccccccccccccc}
				\toprule[0.75pt]
				\multicolumn{1}{l}{\multirow{2}{*}{Surrogate}} & \multirow{2}{*}{Attack} & \multirow{2}{*}{Venue} & \multicolumn{3}{c}{ResNet18}   & \multicolumn{3}{c}{ResNet101}  & \multicolumn{3}{c}{ResNeXt50}  & \multicolumn{3}{c}{DenseNet121} \\ 
				\cmidrule(lr){4-6} \cmidrule(lr){7-9} \cmidrule(lr){10-12} \cmidrule(lr){13-15}
				\multicolumn{1}{l}{}                               &                         &                        & \multicolumn{1}{c}{Ori} & \multicolumn{1}{c}{+Ours} & \multicolumn{1}{c}{$\Delta$} & \multicolumn{1}{c}{Ori} & \multicolumn{1}{c}{+Ours} & \multicolumn{1}{c}{$\Delta$} & \multicolumn{1}{c}{Ori} & \multicolumn{1}{c}{+Ours} & \multicolumn{1}{c}{$\Delta$} & \multicolumn{1}{c}{Ori} & \multicolumn{1}{c}{+Ours} & \multicolumn{1}{c}{$\Delta$} \\ 
				\midrule[0.5pt]
				\multirow{4}{*}{RN18}                        & \scriptsize{MI-FGSM}  & CVPR'18  & 100 & 100 & 0.0 & 40.3  & 47.9  &\textcolor[HTML]{C3375A}{7.6}  & 43.4  & 50.7  &\textcolor[HTML]{C3375A}{7.3}  & 51.0  & 56.6  &\textcolor[HTML]{C3375A}{5.6}  \\
				& AdMix   & ICCV'21  & 100 & 100 & 0.0 & 62.1  & 64.1  &\textcolor[HTML]{C3375A}{2.0}  & 65.1  & 65.2  &\textcolor[HTML]{C3375A}{0.1}  & 73.5  & 73.9  &\textcolor[HTML]{C3375A}{0.4}  \\
				& PAM     & CVPR'23  & 100 & 100 & 0.0 & 45.6  & 53.5  &\textcolor[HTML]{C3375A}{7.9}  & 48.8  & 55.6  &\textcolor[HTML]{C3375A}{6.8}  & 57.7  & 65.6  &\textcolor[HTML]{C3375A}{7.9}  \\
				& BSR     & CVPR'24  & 100 & 100 & 0.0 & 79.6  & 84.4  &\textcolor[HTML]{C3375A}{4.8}  & 81.7  & 86.1  &\textcolor[HTML]{C3375A}{4.4}  & 88.7  & 91.7  &\textcolor[HTML]{C3375A}{3.0}  \\ 
				\midrule[0.5pt]
				\multirow{4}{*}{RN101}                       & \scriptsize{MI-FGSM}  & CVPR'18  & 44.4  & 55.2  &\textcolor[HTML]{C3375A}{10.8}  & 99.6  & 100 &\textcolor[HTML]{C3375A}{0.4}  & 44.7  & 55.9  &\textcolor[HTML]{C3375A}{11.2}  & 44.8  & 53.5  &\textcolor[HTML]{C3375A}{8.7}  \\
				& AdMix   & ICCV'21  & 75.7  & 76.2  &\textcolor[HTML]{C3375A}{0.5}  & 99.7  & 99.7  & 0.0  & 73.0  & 73.6  &\textcolor[HTML]{C3375A}{0.6}  & 75.5  & 76.1  &\textcolor[HTML]{C3375A}{0.6}  \\
				& PAM     & CVPR'23  & 59.3  & 68.6  &\textcolor[HTML]{C3375A}{9.3}  & 99.3  & 98.9  &\textcolor[HTML]{8ab446}{-0.4}  & 55.4  & 63.2  &\textcolor[HTML]{C3375A}{7.8}  & 58.2  & 67.9  &\textcolor[HTML]{C3375A}{9.7}  \\
				& BSR     & CVPR'24  & 87.3  & 91.3  &\textcolor[HTML]{C3375A}{4.0}  & 99.9  & 99.9  & 0.0  & 87.4  & 91.7  &\textcolor[HTML]{C3375A}{4.3}  & 86.1  & 90.4  &\textcolor[HTML]{C3375A}{4.3}  \\ 
				\midrule[0.5pt]
				\multirow{4}{*}{RNX50}                       & \scriptsize{MI-FGSM}  & CVPR'18  & 38.8  & 48.5  &\textcolor[HTML]{C3375A}{9.7}  & 38.4  & 49.4  &\textcolor[HTML]{C3375A}{11.0}  & 99.0  & 100 &\textcolor[HTML]{C3375A}{1.0}  & 41.9  & 51.3  &\textcolor[HTML]{C3375A}{9.4}  \\
				& AdMix   & ICCV'21  & 67.9  & 68.3  &\textcolor[HTML]{C3375A}{0.4}  & 63.1  & 63.8  &\textcolor[HTML]{C3375A}{0.7}  & 98.0  & 99.1  &\textcolor[HTML]{C3375A}{1.1}  & 69.7  & 72.3  &\textcolor[HTML]{C3375A}{2.6}  \\
				& PAM     & CVPR'23  & 52.1  & 58.5  &\textcolor[HTML]{C3375A}{6.4}  & 45.4  & 53.7  &\textcolor[HTML]{C3375A}{8.3}  & 98.2  & 97.6  &\textcolor[HTML]{8ab446}{-0.6}  & 56.1  & 64.2  &\textcolor[HTML]{C3375A}{8.1}  \\
				& BSR     & CVPR'24  & 81.3  & 87.1  &\textcolor[HTML]{C3375A}{5.8}  & 76.6  & 84.8  &\textcolor[HTML]{C3375A}{8.2}  & 99.5  & 100 &\textcolor[HTML]{C3375A}{0.5}  & 83.8  & 90.2  &\textcolor[HTML]{C3375A}{6.4}  \\ 
				\midrule[0.5pt]
				\multirow{4}{*}{DN121}                     & \scriptsize{MI-FGSM}  & CVPR'18  & 51.2  & 59.3  &\textcolor[HTML]{C3375A}{8.1}  & 42.5  & 50.7  &\textcolor[HTML]{C3375A}{8.2}  & 47.3  & 56.4  &\textcolor[HTML]{C3375A}{9.1}  & 99.9  & 99.9  & 0.0  \\
				& AdMix   & ICCV'21  & 78.4  & 76.9  &\textcolor[HTML]{8ab446}{-1.5}  & 68.8  & 68.4  &\textcolor[HTML]{8ab446}{-0.4}  & 74.4  & 73.5  &\textcolor[HTML]{8ab446}{-0.9}  & 99.8  & 100 &\textcolor[HTML]{C3375A}{0.2}  \\
				& PAM     & CVPR'23  & 65.5  & 70.0  &\textcolor[HTML]{C3375A}{4.5}  & 52.6  & 58.1  &\textcolor[HTML]{C3375A}{5.5}  & 59.3  & 63.4  &\textcolor[HTML]{C3375A}{4.1}  & 99.6  & 99.8  &\textcolor[HTML]{C3375A}{0.2}  \\
				& BSR     & CVPR'24  & 88.7  & 92.6  &\textcolor[HTML]{C3375A}{3.9}  & 79.2  & 84.5  &\textcolor[HTML]{C3375A}{5.3}  & 87.2  & 92.0  &\textcolor[HTML]{C3375A}{4.8}  & 100 & 100 & 0.0  \\
				\toprule[0.75pt]
			\end{tabular}
	}}
\end{table}

\subsection{Ablation Study}
We report the average ASR over black-box attacks and the hyper-parameters keep still as in training unless specified.

\textbf{Individual component}. The results are shown in Table~\ref{tbl:abl_component}, where the baseline is vanilla PGD, which is very poor as it neglects temporal cues. When adding the background attack loss (row~2), the performance is doubled; when using the Background Adversarial Mixup (BAM) module (row~3) or Background-induced Temporal Gradient enhancement (BTG) module (row~4), the performance is largely improved and the former is better. When using both of BAM and BTG (Ours), the performance achieves the best. When abandoning the reward $R_{tbc}$ (row~5), the performance degrades, which demonstrates the background consistency along the temporal dimension affects the total reward that decides the selected background frame. When abandoning $\mathcal{L}_{tgc}$ (row~6), the attack performances deteriorate significantly by about 7\% to 12\%, which validates the importance of gradient consistency between two nearby adversarial frames.

\begin{table}[!t]
	\centering
	\caption{Ablation of components on UCF101. ``w/o'' is without.}
	\label{tbl:abl_component}
	\vspace{-2mm}
	\scalebox{0.8}{   
		\setlength{\tabcolsep}{0.95mm}{
			\begin{tabular}{lccccccccccccccccc}
				\toprule[0.75pt]
				\multirow{2}{*}{Attack Method} & \multicolumn{4}{c}{SF/TPN$\rightarrow$NL}          & \multicolumn{4}{c}{NL/TPN$\rightarrow$SF}          & \multicolumn{4}{c}{NL/SF$\rightarrow$TPN}                    \\ 
				\cmidrule(lr){2-5} \cmidrule(lr){6-9} \cmidrule(lr){10-13}
				& & RN101 & RN50 & & & RN101 & RN50 & & & RN101 & RN50 & \\ 
				\midrule[0.5pt]
				Baseline                              &  & 14.85     & 17.33  & &  & 10.89     & 15.35  & &  & 9.41      & 12.87  &  \\
				+$\mathcal{L}_{back}$                  &  & 24.50     & 33.42  & &  & 30.69     & 32.43  & &  & 24.75     & 25.24  &  \\
				+BAM                       &  & 53.96     & 60.40  & &  & 49.01     & 52.97  & &  & 35.15     & 47.52  &  \\
				+BTG &  & 46.53     & 54.95  & &  & 43.56     & 48.02  & &  & 30.69     & 50.00  &  \\
				Ours w/o $R_{tbc}$&  & \underline{65.84}     & \underline{73.26}  & &  & \underline{62.87}     & \underline{60.40}  & &  & \underline{51.98}     & \underline{53.96}  &  \\
				Ours w/o $\mathcal{L}_{tgc}$         &  & 59.90     & 65.35  & &  & 55.94   & 57.43  & &  & 45.05    & 49.50  &  \\
				Ours   & & \textbf{71.29} & \textbf{77.23} & & & \textbf{67.82} & \textbf{68.32} & & & \textbf{52.97} & \textbf{60.40}  & \\ 
				\toprule[0.75pt]
			\end{tabular}
		}
	}
\end{table}

\textbf{$N_{iter}$ and $\epsilon$}. The results are shown in Table~\ref{tbl:abl_iter_epsilon_ucf}, which shows that the attack performance is naturally improved with the increasing iteration number $N_{iter}$ and the maximum perturbation $\epsilon$. However, we observe that when $N_{iter}$ rises from 10 to 20, the performance becomes stable and even suffers from the bottleneck (row~2/4 in group~1) at much larger computational cost. So we choose 10 for $N_{iter}$. Besides, while the performance is greatly boosted by using larger perturbations, \eg, $\epsilon$=32, the adversarial examples are easily found by human, which violates the attack rule that requires small human-imperceptible noise. So we use the trade-off 16 for $\epsilon$.  

\begin{table}[!t]
	\centering
	\caption{Ablation of $N_{iter}$ and $\epsilon$ on UCF101.}
		\vspace{-3mm}
	\label{tbl:abl_iter_epsilon_ucf}
	\scalebox{0.65}{   
		\setlength{\tabcolsep}{1.9mm}{
			\begin{tabular}{llcccccccc}
				\toprule[0.75pt]
				\multirow{2}{*}{Surrogate} & \multirow{2}{*}{Target} & \multicolumn{4}{c}{$N_{iter}$} & \multicolumn{4}{c}{$\epsilon$} \\
				\cmidrule(lr){3-6} \cmidrule(lr){7-10}
				& & 1 & 5 & \textcolor{blue}{10}\textsuperscript{*} & 20 & 4 & 8 & \textcolor{blue}{16}\textsuperscript{*} & 32 \\
				\midrule[0.5pt]
				\multirow{4}{*}{NL-101}&SF-101 & 30.69  & 60.40  & \textcolor{blue}{71.28}  & 73.27  & 2.97  & 33.66  & \textcolor{blue}{71.28}  & 98.02  \\
				&SF-50 & 26.73  & 56.44  & \textcolor{blue}{69.32} & 68.32  & 3.96  & 32.67  & \textcolor{blue}{69.32}  & 98.02  \\
				&TPN-101 & 30.69  & 46.53  & \textcolor{blue}{53.47}  & 56.44  & 0.00  & 17.82  & \textcolor{blue}{53.47}  & 91.09  \\
				&TPN-50 & 27.72  & 48.51  & \textcolor{blue}{54.45}  & 54.46  & 0.00  & 15.84  & \textcolor{blue}{54.45}  & 94.06  \\
				\midrule[0.5pt]
				\multirow{4}{*}{SF-101}
				&NL-101 & 32.67  & 53.47  & \textcolor{blue}{73.27}  & 72.28  & 4.95  & 35.64  & \textcolor{blue}{73.27}  & 95.05  \\
				&NL-50 & 34.65  & 54.46  & \textcolor{blue}{78.22}  & 81.19  & 10.89  & 36.63  & \textcolor{blue}{78.22}  & 100.00  \\
				&TPN-101 & 27.72  & 40.59  & \textcolor{blue}{52.48}  & 54.46  & 2.97  & 24.75  & \textcolor{blue}{52.48}  & 84.16  \\
				&TPN-50 & 30.69  & 49.50  & \textcolor{blue}{66.34}  & 67.33  & 6.93  & 29.70  & \textcolor{blue}{66.34}  & 92.08  \\
				\midrule[0.5pt]
				\multirow{4}{*}{TPN-101}&NL-101 & 29.70  & 65.35  & \textcolor{blue}{69.31}  & 72.28  & 6.93  & 39.60  & \textcolor{blue}{69.31}  & 91.09  \\
				&NL-50 & 34.65  & 64.36  & \textcolor{blue}{76.24}  & 77.23  & 13.86  & 36.63  & \textcolor{blue}{76.24}  & 97.03  \\
				&SF-101 & 34.65  & 51.49  & \textcolor{blue}{64.36} & 65.35  & 5.94  & 33.66  & \textcolor{blue}{64.36}  & 93.07  \\
				&SF-50 & 36.63  & 55.45  & \textcolor{blue}{67.33}  & 68.32  & 4.95  & 37.62  & \textcolor{blue}{67.33}  & 94.06  \\
				\toprule[0.75pt]
			\end{tabular}
		}
	}

\end{table}

\textbf{$\gamma$ and $\beta$}. The results are shown in Table~\ref{tbl:abl_gamma_beta}, where the attack performance rises up when $\gamma$ in the adversarial mixup increases from 0.1 to 0.2 and then decreases when $\gamma$ is over 0.2. This demonstrates that the background frame should not dominate the mixup. Similar observations are found for the hyper-parameter $\beta$ of the temporal gradient consistency loss $\mathcal{L}_{tgc}$, when it starts from 0.01 to 0.1. This suggests that neither too small nor large values are taken for $\beta$. 

\begin{table}[!t]
	\centering
	\caption{Ablation of $\gamma$ (adversarial mixup) and $\beta$ (temporal gradient consistency loss) on UCF101.}
	\vspace{-2mm}
	\label{tbl:abl_gamma_beta}
	\scalebox{0.6}{   
		\setlength{\tabcolsep}{1.9mm} 
		\begin{tabular}{lllllllllll}
			\toprule[0.75pt]
			\multirow{2}{*}{Surrogate}       & \multirow{2}{*}{Target} & \multicolumn{4}{c}{$\gamma$}                                                                             & \multicolumn{5}{c}{$\beta$}                                                                                                          \\ \cmidrule(lr){3-6} \cmidrule(lr){7-11}
			&    & \multicolumn{1}{c}{0.1} & \multicolumn{1}{c}{\textcolor{blue}{0.2}\textsuperscript{*}} & \multicolumn{1}{c}{0.4} & \multicolumn{1}{c}{0.6} & \multicolumn{1}{c}{0.01} & \multicolumn{1}{c}{0.05} & \multicolumn{1}{c}{\textcolor{blue}{0.1}\textsuperscript{*}} & \multicolumn{1}{c}{0.2} & \multicolumn{1}{c}{0.5} \\ 
			\midrule[0.5pt]
			\multirow{4}{*}{NL-101}       & SF-101            & 59.41                   & \textcolor{blue}{71.28}                  & 41.58                   & 37.62                   & 58.42                    & 63.37                    & \textcolor{blue}{71.28}                   & 59.41                   & 48.51                   \\ 
			& SF-50             & 61.39                   & \textcolor{blue}{69.32}                   & 39.60                   & 30.69                   & 57.43                    & 59.41                    & \textcolor{blue}{69.32}                   & 57.43                   & 51.49                   \\ 
			& TPN-101                 & 43.56                   & \textcolor{blue}{53.47}                   & 32.67                   & 32.67                   & 43.56                    & 46.53                    & \textcolor{blue}{53.47}                   & 46.53                   & 32.67                   \\ 
			& TPN-50                  & 41.58                   & \textcolor{blue}{54.45}                   & 29.70                   & 29.70                   & 43.56                    & 42.57                    & \textcolor{blue}{54.45}                   & 40.59                   & 33.66                   \\ 
			\midrule[0.5pt]
			\multirow{4}{*}{SF-101} & NL-101                  & 64.36                   & \textcolor{blue}{73.27}                   & 45.54                   & 37.62                   & 58.42                    & 60.40                    & \textcolor{blue}{73.27}                   & 55.45                   & 43.56                   \\ 
			& NL-50                   & 70.30                   & \textcolor{blue}{78.22}                   & 48.51                   & 38.61                   & 61.39                    & 65.35                    & \textcolor{blue}{78.22}                   & 62.38                   & 48.51                   \\ 
			& TPN-101                 & 42.57                   & \textcolor{blue}{52.48}                   & 34.65                   & 31.68                   & 39.60                    & 47.52                    & \textcolor{blue}{52.48}                   & 43.56                   & 35.64                   \\ 
			& TPN-50                  & 50.50                    & \textcolor{blue}{66.34}                   & 39.60                   & 34.65                   & 51.49                    & 58.42                    & \textcolor{blue}{66.34}                   & 53.47                   & 41.58                   \\ 
			\midrule[0.5pt]
			\multirow{4}{*}{TPN-101}      & NL-101                  & 51.49                   & \textcolor{blue}{69.31}                   & 37.62                   & 36.63                   & 55.45                    & 63.37                    & \textcolor{blue}{69.31}                   & 51.49                   & 43.56                   \\ 
			& NL-50                   & 57.43                   & \textcolor{blue}{76.24}                   & 47.52                   & 40.59                   & 64.36                    & 70.30                    & \textcolor{blue}{76.24}                   & 66.34                   & 56.44                   \\ 
			& SF-101            & 53.47                   & \textcolor{blue}{64.36}                   & 40.59                   & 32.67                   & 53.47                    & 62.38                    & \textcolor{blue}{64.36}                   & 50.50                   & 42.57                   \\ 
			& SF-50             & 51.49                   & \textcolor{blue}{67.33}                   & 42.57                   & 38.61                   & 48.51                    & 59.41                    & \textcolor{blue}{67.33}                   & 44.55                   & 35.64                   \\ 
			\toprule[0.75pt]
		\end{tabular}
	}
\end{table}

\subsection{Computational Efficiency}
We report the parameter size, the computational cost (GFLOPs), the inference speed (fps), and the ASR score on UCF101 and Kinetics-400 in Table~\ref{tbl:compute_cost}. From the table, our BMTC method enjoys a more satisfying tradeoff between performance and efficiency. For example, compared to Temporal Translation (TT) \cite{wei-aaai2022-tt}, ours have much higher ASR score (69.31 vs 52.47 on UCF101, 75.69 vs 62.75 on Kinetics-400) with only nearly one-tenth GFLOPs of TT at a 16 times faster inference speed.

\begin{table}[!t]
	\centering
	\caption{Computational cost comparison.}
		\vspace{-2mm}
	\label{tbl:compute_cost}
	\scalebox{0.8}{   
		\setlength{\tabcolsep}{1.2mm}{
			\begin{tabular}{lrrrr c c}
				\toprule[0.75pt] 
				\multirow{2}{*}{Attack Method} & \multicolumn{1}{c}{Params} & \multicolumn{1}{c}{FLOPs} & \multicolumn{1}{c}{FPS} & & \multicolumn{2}{c}{ASR$\uparrow$}  \\
				\cmidrule{2-4} \cmidrule{6-7}
				& (M)$\downarrow$ &(G)$\downarrow$ &$\uparrow$ & & UCF101 & Kinetics-400 \\
				\midrule[0.5pt]
				PGD \scriptsize{\cite{madry-iclr2018-pgd}}  & 99.7            & 217.4               & 357.2     &               & 11.88       &  15.54             \\
				AA \scriptsize{\cite{croce-icml2020-aa}}    & 99.7            & 295.3               & 290.4 &                   & 15.35       &  22.96              \\
				TT \scriptsize{\cite{wei-aaai2022-tt}}      & 99.7            & 3634.1              & 17.8  &                   & \underline{52.47}       &  \underline{62.75}              \\
				Ours                                        & 123.4           & 391.6               & 288.9  &                  & \textbf{69.31}       & \textbf{75.69}      \\ 
				\toprule[0.75pt]                    
			\end{tabular}%
		}
	}
\end{table}
 
\subsection{Visualization of Adversarial Examples}
\begin{figure}[!t]
	\centering
	\includegraphics[width=0.49\linewidth]{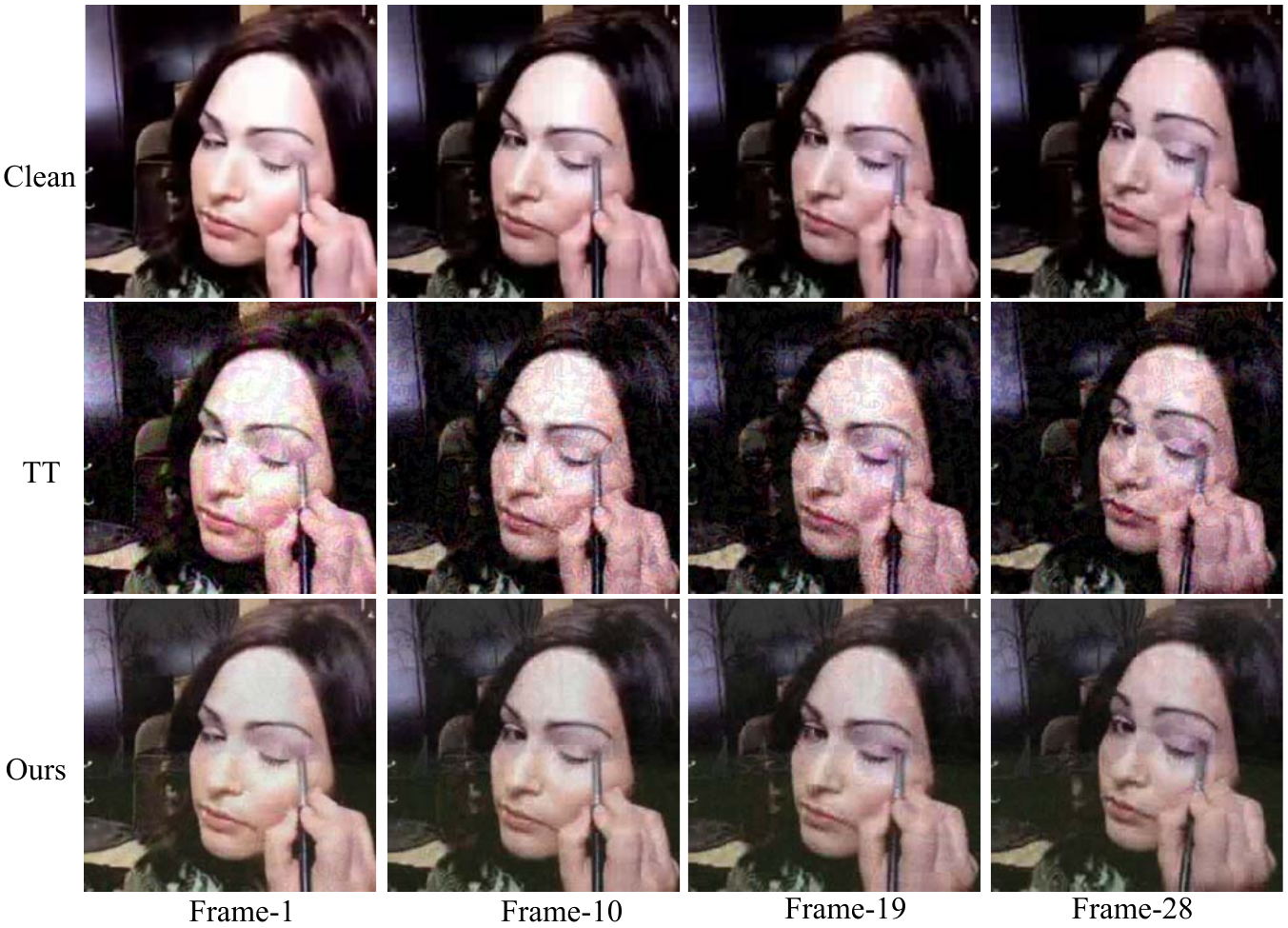} 
	\includegraphics[width=0.49\linewidth]{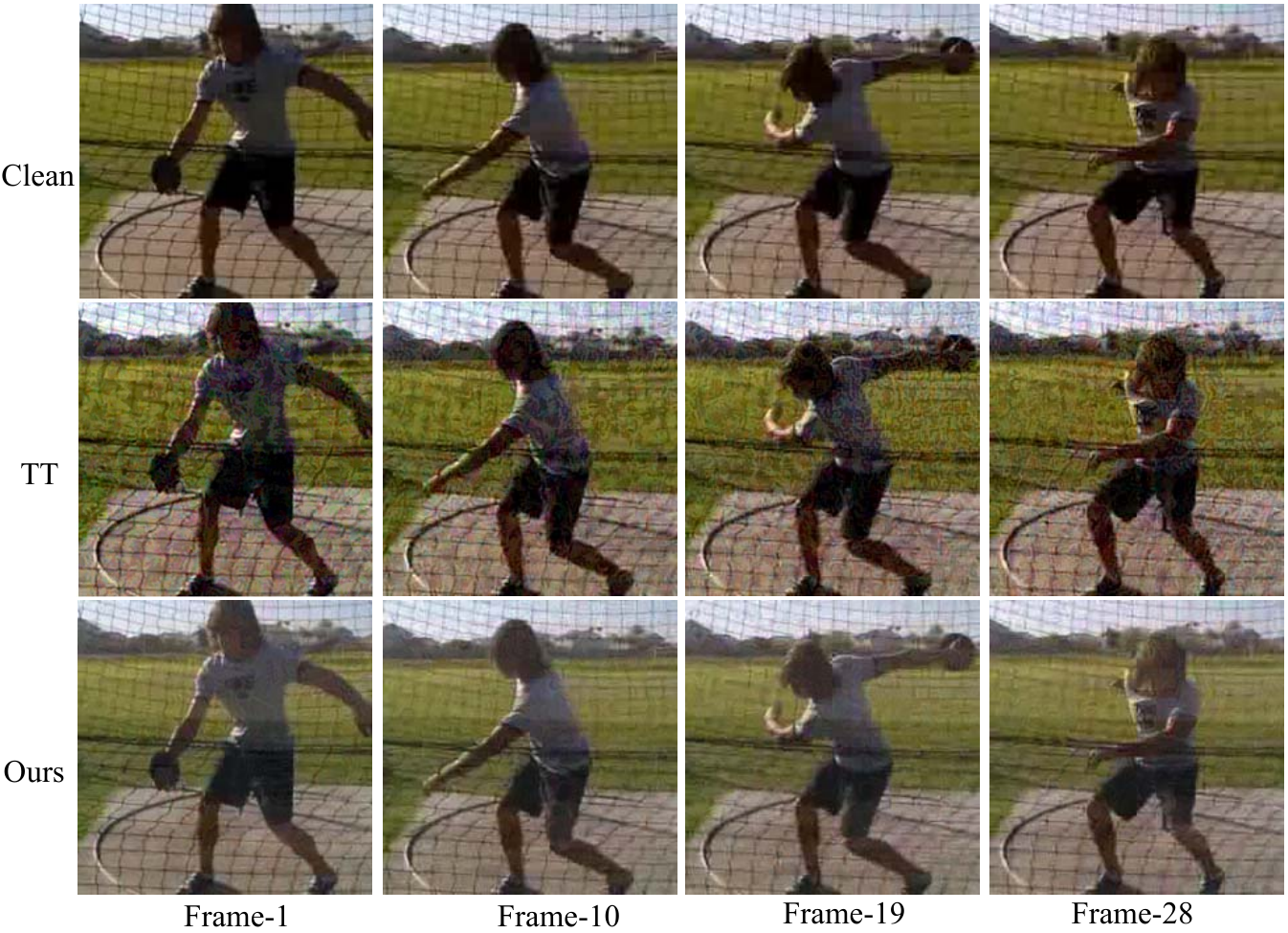}
	\caption{Examples of UCF101 (left) and Kinetics-400 (right).}
	\label{fig:vis_ucf_kinetics}
	\vspace{-2mm}
\end{figure}

To visualize the performance of our attack, we randomly chose one video from UCF101 and Kinetics-400, respectively, and show their adversarial examples in Fig.~\ref{fig:vis_ucf_kinetics} (zoom in for better view). Compared to the baseline Temporal Translation (TT) \cite{wei-aaai2022-tt}, the perturbations on the adversarial examples of ours are almost imperceivable by human but the model can be fooled to make wrong predictions.

\section{Conclusion}
\label{sec:conclusion}
This work presents a transferable black-box adversarial attack method for action recognition by considering both the temporal background consistency and the temporal gradient consistency. In particular, we adopt the adversarial mixup strategy to mix the clean sample with the background frame from other categories. To ensure the attack ability of the background frame, we design a reward function that considers the temporal consistency among nearby frames and the transferability across different models. Moreover, we strengthen the transferability of adversarial example by adopting the background-induced temporal consistency on the gradients of sample across frames. Empirical studies on both video and image datasets validate the effectiveness of the proposed attack on several models with different architectures.

\newpage

\appendix

\noindent\textbf{\Large {Appendix: Temporal Consistency Constrained Transferable Adversarial Attacks with Background Mixup for Action Recognition}}\\

This supplement provides the detailed related works, the descriptions of datasets and compared methods, as well as the ablations of $N_{iter}$ and $\epsilon$ on ImageNet. In addition, we make an analysis on the varying gains on ImageNet.

To facilitate reading the manuscript, our code is available at {\b \href{https://github.com/mlvccn/BMTC\_TransferAttackVid}{https://github.com/mlvccn/BMTC\_TransferAttackVid}}.

\section{Related Works}
This section reviews typical transfer-based attack methods in both image domain and video domain, and shows some action recognition methods closely related to our work.

\subsection{Transfer-based Image Attacks}
To generate adversarial examples with high transferability, previous methods often adopt white-box attacks, such as Fast Gradient Sign Method (FGSM) \cite{goodefellow-iclr2015-fgsm} that updates the gradient on the clean sample to maximize the loss function, Basic Iterative Method (BIM) \cite{kurakin-arxiv2016-adversarial} that iteratively applies FGSM multiple times, and Projected Gradient Descent (PGD) \cite{madry-iclr2018-pgd} that is also an iterative FGSM. To improve the transferability of adversarial examples on black-box models, there are three strategies: 1) \textit{data augmentation} (\ie,input transformation), \eg, diversity input attack \cite{xie-cvpr2019-dim}, Scale-Invariant Method (SIM) \cite{lin-iclr2020-sini}, Translation-Invariant (TI) attack \cite{dong-cvpr2019-tim}, Adversarial Mixup (AdMix) \cite{wang-iccv2021-admix} that mixes two images using Mixup \cite{zhang-iclr2018-mixup}, Adaptive Image Transformation (AIT) \cite{yuan-eccv2022-ait}, Path-Augmented Method (PAM) \cite{wang-cvpr2023-pam} that constructs a candidate augmentation path pool for path selection with greedy search, and Learning to Transform (L2T) \cite{zhu-cvpr2024-transfer} that increases the diversity of transformed images by selecting the optimal combination of input transformations using reinforcement learning; 2) \textit{gradient modification}, \eg, Momentum Iterative (IM) attack \cite{dong-cvpr2018-mifgsm}, Nesterov Accelerated Gradient (NAG) \cite{lin-iclr2020-sini}, Skip Gradient Method (SGM) \cite{wu-iclr2020-sgm}, Variance Momentum Iterative tuning (VMI) \cite{wang-cvpr2021-vmi}, Gradient-Related Adversarial Attack (GRA) \cite{zhu-iccv2023-gra} that considers the gradient sign change of adversarial perturbation and leverages the gradient correlation to alleviate the sign fluctuation problem, and Backward Propagation Attack (BPA) \cite{wang-nips2024-backward} that incorporates a temperature parameter in softmax function to smooth the derivatives of max-pooling layers for reducing gradient truncation; 3) \textit{feature disruption}, \eg, Intermediate Level Attack (ILA) \cite{huang-iccv2019-ila}, Attention-guided Transfer Attack (ATA) \cite{wu-cvpr2020-ata}, Feature Importance-Aware attack (FIA) \cite{wang-iccv2021-fia}, Neuron Attribution-based Attack (NAA) \cite{zhang-cvpr2022-naa} that conducts feature-level attacks with more accurate neuron importance estimations, and ILA with Data Augmentation (ILA-DA) \cite{yan-iclr2023-ilada} that applies automated data augmentation, reverse adversarial update, and attack interpolation techniques onto ILA to strengthen the attack transferability. These methods may provide inspiration for video attacks.

\subsection{Transfer-based Video Attacks}
Transferable attacks on video models are less explored compared to that on image models. Existing methods attempt to break the temporal relations, \eg, from the gradient update perspective, Temporal Translation (TT) \cite{wei-aaai2022-tt} does multiple translations on frames along the temporal dimension to generate the clips under different translation patterns and optimizes the perturbations on these clips; from the feature disruption perspective, some methods minimize the cosine similarities between the intermediate features of the clean frames \cite{wei-tpami2024-i2v} or the warped clean frames \cite{chen-acmmm2023-gcma} and the adversarial counterparts. Nevertheless, they treat video as a collection of unordered images, ignoring the inherent temporal continuity. In contrast, \cite{wang-aaai2023-i2v} disrupts the temporal local correlations of nearby frames by reducing the similarity of the adversarial counterparts, and minimizes the cosine similarities among the intermediate features of the benign points lying on a convex hull, which leads to diverse adversarial frames. This may incur gradient oscillation, \eg, some gradients update in reverse direction, because the uncertain attack directions vary greatly. 

\subsection{Action Recognition Models}
Much progress has been made in deep neural network based action recognition models, whose security concerns become increasingly important due to the analysis demand on the fast-growing videos collected from numerous surveillance cameras and mobile sensing devices. Existing methods can be generally divided into two groups: 1) Convolutional Neural Networks (CNNs) methods, \eg, Inflated 3D ConvNet (I3D) \cite{carreira-cvpr2017-kinetics}, Non-Local neural networks (NL) \cite{wang-cvpr2018-nl}, SlowFast \cite{feichtenhofer-iccv2019-slowfast} that captures spatial semantics at low frame rate and the motion dynamics at fine temporal resolution, Temporal Pyramid Networks (TPN) \cite{yang-cvpr2020-tpn} that uses the source features and the fusion features to form a feature hierarchy to capture action instances at various tempos; 2) Vision Transformer (ViT) methods, \eg, VideoTransformer Network (VTN) \cite{neimark-cvpr2021-videotransformer} that is built on top of any given 2D spatial network and attends to the entire video, Time-Space Transformer (TimeSformer) \cite{bertasius-icml2021-video} that leverages 3D self-attention over the space-time volume to capture long-range temporal dependency among frames, ViViT (Video Vision Transformer) \cite{arnab-iccv2021-vivit} that extracts spatiotemporal video tokens to be encoded by a series of transformer layers, Motionformer \cite{patrick-nips2021-motionformer} aggregates implicitly motion path information to model the temporal dynamic scenes, Multiscale ViT \cite{fan-iccv2021-mvit} that has multiple feature channel-resolution scale stages which hierarchically expand the channel capacity while reducing the spatial resolution, and VideoSwin \cite{liu-cvpr2022-videoswin} that encourages an inductive bias of locality by adapting the Swin Transformer \cite{liu-iccv2021-swin} for a better speed-accuracy trade-off.
\\

\begin{table}[!t]
	\centering
	\caption{Statistics of data.}
	\vspace{-2mm}
	\label{tbl:dataset}
	\scalebox{1.0}{ 
		\setlength{\tabcolsep}{2mm}{
			\begin{tabular}{ l  r  r  r  r }
				\toprule[0.75pt]
				Dataset  & Train & Val & Test  & $K$\\
				\midrule[0.5pt]
				UCF101      & 9,537  & -  & 3,783  & 101 \\
				Kinetics-400    & 240,436 & 19,877 & 38,733  & 400\\
				ImageNet   & 1,000,000 & 50,000 & 100,000  & 1,000\\
				\toprule[0.75pt]
			\end{tabular}
		}
	}
\end{table}

\section{Experiment Supplement}
\subsection{Datasets}
We conduct experiments on two video benchmarks including UCF101 \cite{soomro-arxiv2012-ucf101} and Kinetics-400  \cite{carreira-cvpr2017-kinetics}, and one image benchmark ImageNet \cite{deng-cvpr2009-imagenet}. Statistics are in Table~\ref{tbl:dataset}. 

\textbf{UCF101} \cite{soomro-arxiv2012-ucf101}\footnote{\url{https://www.crcv.ucf.edu/research/data-sets/ucf101/}} is a widely used dataset for video action recognition, containing 13,320 video clips across 101 action categories. The action categories are diverse, including sports, musical instrument playing, and other scenes. The dataset has approximately 9,537 videos for training and 3,783 videos for testing in each fold. No independent validation set is provided.

\textbf{Kinetics-400} \cite{carreira-cvpr2017-kinetics}\footnote{\url{https://deepmind.com/research/open-source/kinetics}} is a large-scale video action recognition dataset containing around 306,245 videos across 400 distinct action categories. The video clips are sourced from YouTube, covering a wide variety of actions. It has 240,436 videos for training, 19,877 videos for validation, and 38,733 videos for testing.

\textbf{ImageNet} \cite{deng-cvpr2009-imagenet}\footnote{\url{https://www.image-net.org/challenges/LSVRC/index.php}} is a large-scale image classification dataset containing approximately 1,200,000 images across 1,000 different categories. The images are sourced from various platforms, including web scraping and manual annotations. It has 1,000,000 images for training, 50,000 images for validation, and 100,000 images for testing. The wide variety of categories and high-quality annotations make it one of the standard datasets for pretraining and evaluating deep learning models.

\subsection{Compared Methods}
For action recognition, we compare our method with several transfer-based attack baselines including Temporal Translation (TT) \cite{wei-aaai2022-tt}, I2V \cite{wei-cvpr2022-i2v}, Generative Cross-Modal Adversary (GCMA)  \cite{chen-acmmm2023-gcma}, Global-local Characteristic Excited Cross-modal attack (GCEC) \cite{wang-aaai2023-i2v}, and I2V-MF \cite{wei-tpami2024-i2v}. For image classification, we compare our method with Momentum Iterative Fast Gradient Sign Method (MI-FGSM) \cite{dong-cvpr2018-mifgsm}, Adversarial Mixup (AdMix) \cite{wang-iccv2021-admix}, Path-Augmented Method (PAM) \cite{wang-cvpr2023-pam} that is also an adversarial mixup method, and Block Shuffle Rotation (BSR) \cite{wang-cvpr2024-bsr} which is a non-mixup input transformation method.

\subsection{Ablations of $\alpha_1$ and $\alpha_2$ on UCF101}
To examine the impacts of the transfer reward and the temporal background consistency reward, we vary the hyper-parameters $\alpha_1$ and $\alpha_2$ from the sets $\{0, 0.1, 0.3, 0.5\}$ and $\{0, 0.05, 0.1, 0.2\}$, respectively. The results are shown in Table~\ref{tbl:abl_alpha}. From the records, we see that the transfer reward $R_{transfer}$ achieves the best when $\alpha_1$ takes 0.3 while the temporal consistency reward $R_{tbc}$ performs the best when $\alpha_2$ takes 0.1. This indicates that the former contributes more greatly than the latter to the total reward. 

\begin{table}[!t]
	\centering
	\caption{Ablation of $\alpha_1$ and $\alpha_2$ on UCF101.}
	\label{tbl:abl_alpha}
	\scalebox{0.75}{   
		\setlength{\tabcolsep}{0.6mm}{
			\begin{tabular}{llcccccccc}
				\toprule[0.75pt]
				\multicolumn{1}{c}{\multirow{2}{*}{Surrogate}} & \multicolumn{1}{c}{\multirow{2}{*}{Target}} & \multicolumn{4}{c}{$\alpha_1$} & \multicolumn{4}{c}{$\alpha_2$} \\ \cmidrule(lr){3-6} \cmidrule(lr){7-10}
				\multicolumn{1}{c}{} & \multicolumn{1}{c}{} & 0 & 0.1 & 0.3\textsuperscript{*} & 0.5 & 0 & 0.05 & 0.1\textsuperscript{*} & 0.2 \\ \midrule[0.5pt]
				\multirow{4}{*}{NL\_RN101} & SF\_RN101 & 62.38 & 65.35 & \textbf{71.28} & 70.30 & 65.35 & 68.32 & \textbf{71.28} & 69.31 \\
				& SF\_RN50 & 59.41 & 65.35 & \textbf{69.32} & 67.33 & 62.38 & 67.33 & \textbf{69.32} & 68.32 \\
				& TPN\_RN101 & 49.50 & 51.49 & \textbf{53.47} & 53.47 & 52.48 & 52.48 & \textbf{53.47} & 53.47 \\
				& TPN\_RN50 & 46.53 & 50.50 & \textbf{54.45} & 52.48 & 49.50 & 52.48 & \textbf{54.45} & 53.47 \\ \midrule[0.5pt]
				\multirow{4}{*}{SF\_RN101} & NL\_RN101 & 64.36 & 69.31 & \textbf{73.27} & 71.29 & 66.34 & 70.30 & \textbf{73.27} & 72.28 \\
				& NL\_RN50 & 68.32 & 72.28 & \textbf{78.22} & 75.25 & 74.26 & 77.23 & \textbf{78.22} & 76.24 \\
				& TPN\_RN101 & 48.51 & 50.50 & \textbf{52.48} & 49.50 & 50.50 & 51.49 & \textbf{52.48} & 51.49 \\
				& TPN\_RN50 & 53.47 & 64.36 & \textbf{66.34} & 63.37 & 57.43 & 65.35 & \textbf{66.34} & 64.36 \\ \midrule[0.5pt]
				\multirow{4}{*}{TPN\_RN101} & NL\_RN101 & 61.39 & 66.34 & \textbf{69.31} & 67.33 & 64.36 & 67.33 & \textbf{69.31} & 66.34 \\
				& NL\_RN50 & 67.33 & 71.29 & \textbf{76.24} & 73.27 & 71.29 & 75.25 & \textbf{76.24} & 74.26 \\
				& SF\_RN101 & 57.43 & 60.40 & \textbf{64.36} & 63.37 & 59.41 & 62.38 & \textbf{64.36} & 63.37 \\
				& SF\_RN50 & 58.42 & 64.36 & \textbf{67.33} & 65.35 & 61.39 & 65.35 & \textbf{67.33} & 64.36 \\ \toprule[0.75pt]
			\end{tabular}%
		}
	}
\end{table}

\begin{table*}[!t]
	\caption{Ablation of $N_{iter}$ on ImageNet.}
	\label{tbl:abl_iter_imagenet}
	\centering
		\setlength{\tabcolsep}{0.85mm}{
			\begin{tabular}{lllcccccccccccr}
				\toprule[0.75pt]
				& \multicolumn{1}{c}{}                                & \multicolumn{1}{c}{}                        & \multicolumn{12}{c}{Iteration Number $N_{iter}$} \\ 
				\cmidrule(lr){4-15} 
				& \multicolumn{1}{c}{}                                & \multicolumn{1}{c}{}                        & \multicolumn{3}{c}{1}                                                       & \multicolumn{3}{c}{5}                                                       & \multicolumn{3}{c}{10\textsuperscript{*}}                                                      & \multicolumn{3}{c}{20}                                                                \\ 
				\cmidrule(lr){4-6} \cmidrule(lr){7-9} \cmidrule(lr){10-12} \cmidrule(lr){13-15} 
				\multirow{-3}{*}{Surrogate} & \multicolumn{1}{c}{\multirow{-3}{*}{Attack}} & \multicolumn{1}{c}{\multirow{-3}{*}{Venue}} & \multicolumn{1}{c}{Ori} & \multicolumn{1}{c}{+Ours} & \multicolumn{1}{c}{$\Delta$}       & \multicolumn{1}{c}{Ori} & \multicolumn{1}{c}{+Ours} & \multicolumn{1}{c}{$\Delta$}       & \multicolumn{1}{c}{Ori} & \multicolumn{1}{c}{+Ours} & \multicolumn{1}{c}{$\Delta$}        & \multicolumn{1}{c}{Ori} & \multicolumn{1}{c}{+Ours} & \multicolumn{1}{c}{$\Delta$}        \\ 
				\midrule[0.5pt]
				& MIFGSM                                              & CVPR'18                                     & 34.63                   & 40.76                        &\textcolor[HTML]{C3375A}{6.13} & 50.37                   & 56.45                        &\textcolor[HTML]{C3375A}{6.08} & 58.68                   & 63.80                        &\textcolor[HTML]{C3375A}{5.13}  & 58.85              & 66.70                   &\textcolor[HTML]{C3375A}{7.85}  \\
				& AdMix                                              & ICCV'21                                     & 45.42                   & 46.37                        &\textcolor[HTML]{C3375A}{0.95} & 70.20                   & 70.99                        &\textcolor[HTML]{C3375A}{0.79} & 75.18                   & 75.80                        &\textcolor[HTML]{C3375A}{0.63}  & 77.98                & 77.64                    &\textcolor[HTML]{8ab446}{-0.34} \\
				& PAM                                                 & CVPR'23                                     & 41.51                   & 46.04                        &\textcolor[HTML]{C3375A}{4.53} & 58.71                   & 63.57                        &\textcolor[HTML]{C3375A}{4.86} & 63.03                   & 68.68                        &\textcolor[HTML]{C3375A}{5.65}  & 64.59                & 70.25                     &\textcolor[HTML]{C3375A}{5.66}  \\
				\multirow{-4}{*}{ResNet18}           & BSR                                                 & CVPR'24                                     & 60.71                   & 64.33                        &\textcolor[HTML]{C3375A}{3.63} & 82.62                   & 87.19                        &\textcolor[HTML]{C3375A}{4.56} & 87.50                   & 90.55                        &\textcolor[HTML]{C3375A}{3.05}  & 88.59                & 93.00                    &\textcolor[HTML]{C3375A}{4.41}  \\ 
				\midrule[0.5pt]
				& MIFGSM                                              & CVPR'18                                     & 34.16                   & 43.81                        &\textcolor[HTML]{C3375A}{9.65} & 48.78                   & 58.25                        &\textcolor[HTML]{C3375A}{9.47} & 58.38                   & 66.15                        &\textcolor[HTML]{C3375A}{7.78}  & 59.23     & 66.05                   &\textcolor[HTML]{C3375A}{6.82}  \\
				& AdMix                                              & ICCV'21                                     & 49.16                   & 51.33                        &\textcolor[HTML]{C3375A}{2.18} & 77.00                   & 78.64                        &\textcolor[HTML]{C3375A}{1.64} & 80.98                   & 81.40                        &\textcolor[HTML]{C3375A}{0.43}  & 80.79              & 84.04                &\textcolor[HTML]{C3375A}{3.25}  \\
				& PAM                                                 & CVPR'23                                     & 42.23                   & 47.24                        &\textcolor[HTML]{C3375A}{5.01} & 64.56                   & 69.17                        &\textcolor[HTML]{C3375A}{4.61} & 68.05                   & 74.65                        &\textcolor[HTML]{C3375A}{6.60}  & 68.46           & 78.42                 &\textcolor[HTML]{C3375A}{9.96}  \\
				\multirow{-4}{*}{ResNet101}          & BSR                                                 & CVPR'24                                     & 66.83                   & 70.71                        &\textcolor[HTML]{C3375A}{3.88} & 85.20                   & 89.05                        &\textcolor[HTML]{C3375A}{3.85} & 90.18                   & 93.33                        &\textcolor[HTML]{C3375A}{3.15}  & 89.98                & 92.93                 &\textcolor[HTML]{C3375A}{2.95}  \\ 
				\midrule[0.5pt]
				& MIFGSM                                             & CVPR'18                                     & 31.74                   & 41.04                        &\textcolor[HTML]{C3375A}{9.30} & 48.77                   & 57.13                        &\textcolor[HTML]{C3375A}{8.36} & 54.53                   & 62.30                        &\textcolor[HTML]{C3375A}{7.78}  & 56.55                & 61.72                     &\textcolor[HTML]{C3375A}{5.17}  \\
				& AdMix                                              & ICCV'21                                     & 44.05                   & 47.31                        &\textcolor[HTML]{C3375A}{3.26} & 70.62                   & 74.01                        &\textcolor[HTML]{C3375A}{3.38} & 74.68                   & 75.88                        &\textcolor[HTML]{C3375A}{1.20}  & 77.53                & 79.85                     &\textcolor[HTML]{C3375A}{2.32}  \\
				& PAM                                                 & CVPR'23                                     & 37.36                   & 43.49                        &\textcolor[HTML]{C3375A}{6.13} & 58.01                   & 63.44                        &\textcolor[HTML]{C3375A}{5.43} & 62.95                   & 68.50                        &\textcolor[HTML]{C3375A}{5.55}  & 63.35                & 67.64                     &\textcolor[HTML]{C3375A}{4.29}  \\
				\multirow{-4}{*}{ResNeXt50}          & BSR                                                 & CVPR'24                                     & 60.55                   & 66.49                        &\textcolor[HTML]{C3375A}{5.94} & 82.01                   & 87.59                        &\textcolor[HTML]{C3375A}{5.58} & 85.30                   & 90.53                        &\textcolor[HTML]{C3375A}{5.22}  & 86.82                & 93.35                     &\textcolor[HTML]{C3375A}{6.53}  \\ 
				\midrule[0.5pt]
				& MIFGSM                                           & CVPR'18                                     & 38.63                   & 44.80                        &\textcolor[HTML]{C3375A}{6.16} & 51.19                   & 56.69                        &\textcolor[HTML]{C3375A}{5.50} & 60.23                   & 66.58                        &\textcolor[HTML]{C3375A}{6.35}  & 63.63                & 65.78                     &\textcolor[HTML]{C3375A}{2.15}  \\
				& AdMix                                              & ICCV'21                                     & 50.32                   & 51.23                        &\textcolor[HTML]{C3375A}{0.90} & 76.27                   & 77.07                        &\textcolor[HTML]{C3375A}{0.81} & 80.35                   & 79.70                        &\textcolor[HTML]{8ab446}{-0.65} & 83.18                & 82.85                     &\textcolor[HTML]{8ab446}{-0.33} \\
				& PAM                                                & CVPR'23                                     & 46.06                   & 50.40                        &\textcolor[HTML]{C3375A}{4.34} & 65.81                   & 70.92                        &\textcolor[HTML]{C3375A}{5.11} & 69.25                   & 72.83                        &\textcolor[HTML]{C3375A}{3.58}  & 69.63                & 74.98                     &\textcolor[HTML]{C3375A}{5.35}  \\
				\multirow{-4}{*}{DenseNet121}        & BSR                                                & CVPR'24                                     & 66.36                   & 69.79                        &\textcolor[HTML]{C3375A}{3.43} & 85.73                   & 89.77                        &\textcolor[HTML]{C3375A}{4.04} & 88.78                   & 92.28                        &\textcolor[HTML]{C3375A}{3.50}  & 92.53                & 93.75                     &\textcolor[HTML]{C3375A}{1.22}  \\ 
				\toprule[0.5pt]
			\end{tabular}%
		}
\end{table*}

\begin{table*}[!t]
	\caption{Ablation of $\epsilon$ on ImageNet.}
	\label{tbl:abl_epsilon_imagenet}
	\centering
			\begin{tabular}{lllccccccccc}
				\toprule[0.75pt]
				& \multicolumn{1}{c}{}                                & \multicolumn{1}{c}{}                        & \multicolumn{9}{c}{Maximum Perturbation $\epsilon$}                                                                                                                   \\ 
				\cmidrule(lr){4-12} 
				& \multicolumn{1}{c}{}                                & \multicolumn{1}{c}{}                        & \multicolumn{3}{c}{4}                           & \multicolumn{3}{c}{8\textsuperscript{*}}                       & \multicolumn{3}{c}{16}                          \\ \cmidrule(lr){4-6} \cmidrule(lr){7-9} \cmidrule(lr){10-12}
				\multirow{-3}{*}{Surrogate} & \multicolumn{1}{c}{\multirow{-3}{*}{Attack}} & \multicolumn{1}{c}{\multirow{-3}{*}{Venue}} & Ori    & +Ours & $\Delta$                           & Ori   & +Ours & $\Delta$                           & Ori    & +Ours & $\Delta$                           \\ 
				\midrule[0.5pt]
				& MIFGSM                                              & CVPR'18                                     & 41.93 & 43.60    &\textcolor[HTML]{C3375A}{1.67} & 58.68 & 63.80    &\textcolor[HTML]{C3375A}{5.13}  & 84.03  & 88.99    &\textcolor[HTML]{C3375A}{4.96} \\
				& AdMix                                              & ICCV'21                                     & 46.98  & 47.85    &\textcolor[HTML]{C3375A}{0.87} & 75.18 & 75.80    &\textcolor[HTML]{C3375A}{0.63}  & 92.85  & 93.40    &\textcolor[HTML]{C3375A}{0.55} \\
				& PAM                                                 & CVPR'23                                     & 42.53  & 43.28    &\textcolor[HTML]{C3375A}{0.75} & 63.03 & 68.68    &\textcolor[HTML]{C3375A}{5.65}  & 86.90  & 90.03    &\textcolor[HTML]{C3375A}{3.13} \\
				\multirow{-4}{*}{ResNet18}           & BSR                                                 & CVPR'24                                     & 62.33  & 63.48    &\textcolor[HTML]{C3375A}{1.15} & 87.50 & 90.55    &\textcolor[HTML]{C3375A}{3.05}  & 98.13  & 99.90    &\textcolor[HTML]{C3375A}{1.78} \\ 
				\midrule[0.5pt]
				& MIFGSM                                              & CVPR'18                                     & 42.08  & 45.98    &\textcolor[HTML]{C3375A}{3.90} & 58.38 & 66.15    &\textcolor[HTML]{C3375A}{7.78}  & 82.08  & 88.22    &\textcolor[HTML]{C3375A}{6.14} \\
				& AdMix                                               & ICCV'21                                     & 50.23  & 51.58    &\textcolor[HTML]{C3375A}{1.35} & 80.98 & 81.40    &\textcolor[HTML]{C3375A}{0.43}  & 95.45  & 95.58    &\textcolor[HTML]{C3375A}{0.13} \\
				& PAM                                                & CVPR'23                                     & 44.13  & 45.93    &\textcolor[HTML]{C3375A}{1.80} & 68.05 & 74.65    &\textcolor[HTML]{C3375A}{6.60}  & 87.82  & 92.50    &\textcolor[HTML]{C3375A}{4.68} \\
				\multirow{-4}{*}{ResNet101}          & BSR                                                 & CVPR'24                                     & 66.95  & 68.75    &\textcolor[HTML]{C3375A}{1.80} & 90.18 & 93.33    &\textcolor[HTML]{C3375A}{3.15}  & 98.83  & 99.70    &\textcolor[HTML]{C3375A}{0.88} \\ 
				\midrule[0.5pt]
				& MIFGSM                                              & CVPR'18                                     & 39.43  & 43.03    &\textcolor[HTML]{C3375A}{3.60} & 54.53 & 62.30    &\textcolor[HTML]{C3375A}{7.78}  & 85.43  & 91.67    &\textcolor[HTML]{C3375A}{6.24} \\
				& AdMix                                             & ICCV'21                                     & 44.83  & 47.80    &\textcolor[HTML]{C3375A}{2.97} & 74.68 & 75.88    &\textcolor[HTML]{C3375A}{1.20}  & 92.18  & 92.96    &\textcolor[HTML]{C3375A}{0.78} \\
				& PAM                                                & CVPR'23                                     & 39.05  & 41.68    &\textcolor[HTML]{C3375A}{2.63} & 62.95 & 68.50    &\textcolor[HTML]{C3375A}{5.55}  & 89.93  & 94.80    &\textcolor[HTML]{C3375A}{4.87} \\
				\multirow{-4}{*}{ResNeXt50}          & BSR                                                & CVPR'24                                     & 61.88  & 65.28    &\textcolor[HTML]{C3375A}{3.40} & 85.30 & 90.53    &\textcolor[HTML]{C3375A}{5.22}  & 96.95  & 98.18    &\textcolor[HTML]{C3375A}{1.23} \\ 
				\midrule[0.5pt]
				& MIFGSM                                             & CVPR'18                                     & 44.35  & 45.78    &\textcolor[HTML]{C3375A}{1.43} & 60.23 & 66.58    &\textcolor[HTML]{C3375A}{6.35}  & 81.75  & 87.04    &\textcolor[HTML]{C3375A}{5.29} \\
				& AdMix                                               & ICCV'21                                     & 52.08  & 52.63    &\textcolor[HTML]{C3375A}{0.55} & 80.35 & 79.70    &\textcolor[HTML]{8ab446}{-0.65} & 95.075 & 95.36    &\textcolor[HTML]{C3375A}{0.28} \\
				& PAM                                                 & CVPR'23                                     & 46.30  & 47.03    &\textcolor[HTML]{C3375A}{0.73} & 69.25 & 72.83    &\textcolor[HTML]{C3375A}{3.58}  & 87.65  & 90.21    &\textcolor[HTML]{C3375A}{2.56} \\
				\multirow{-4}{*}{DenseNet121}        & BSR                                                 & CVPR'24                                     & 66.45  & 67.60    &\textcolor[HTML]{C3375A}{1.15} & 88.78 & 92.28    &\textcolor[HTML]{C3375A}{3.50}  & 98.175 & 99.90    &\textcolor[HTML]{C3375A}{1.73} \\ 
				\toprule[0.75pt]
			\end{tabular}%
\end{table*}

\subsection{Ablations of $N_{iter}$ and $\epsilon$ on ImageNet}
The results of iteration number $N_{iter}$ are shown in Table~\ref{tbl:abl_iter_imagenet}. Here, ``Ori'' denotes the vanilla attack method (column~2). From the table, the attack performance is gradually improved as iteration number $N_{iter}$ increases. But the increasing tendency becomes smooth when it surpasses 10. For AdMix \cite{wang-iccv2021-admix}, the performance improvement is marginal or even worse, which might be the reason that our method also adopt adversarial mixup and the temporal loss is unused in image domain. 

The results of maximum perturbation $\epsilon$ are shown in Table~\ref{tbl:abl_epsilon_imagenet}. From the table, we observe that the attack performance is upgraded when $\epsilon$ begins with 4 and degenerated after 8. When $\epsilon$ takes 16, the adversarial examples are more easily found by human that that takes 8. So we take the sensible 8 for $\epsilon$ in image classification task.

\subsection{Discussion on the varying gains on ImageNet}
When we carefully examine the result tables of ImageNet, it will appear some negative gains. To clarify this issue, we make the following analysis. We adopt the gradient-based MI-FGSM \cite{dong-cvpr2018-mifgsm}, the mixup-based AdMix \cite{wang-iccv2021-admix} and PAM \cite{wang-cvpr2023-pam}, and the non-mixup input transformation method BSR \cite{wang-cvpr2024-bsr}. The coupling performance is the best when combining ours with the gradient-based attack method  MI-FGSM, due to their strong complementary property. The coupling performance is the degenerated when combining ours with the non-mixup method BSR and the pure-color mixup method PAM, because their complementary information is reduced. The coupling performance is the weakest when combining ours with the AdMix that adopts the most similar mixup method with ours. 

\subsection{Cross-model attack on Kinetics-400}
To examine the attack ability of the proposed method on the target models with different architectures, we employ CNN-based surrogate model to attack three models, including VideoMamba \cite{li-eccv2024-videomamba}, InternVideo2 \cite{wang-eccv2024-internvideo2}, and InternVideo2.5 \cite{wang-arxiv2025-internvideo25}. Note that Mamba \cite{gu-arxiv2024-mamba} is a sequential modeling architecture, which exhibits the strong representation ability in an efficient way, and Li \etal~\cite{li-eccv2024-videomamba} introduce Mamba to the action recognition model to alleviate the problem of local redundancy and global temporal modeling. The latter two adopt ViT as the backbone, and consist of video encoder, multimodal alignment module, and language model. 

The cross-model attack results on Kinetics-400 are shown in Table~\ref{tbl:cross_attack_kinetics}, from which we see that our method consistently outperforms the competitive alternatives. The superiority of our method can be attributed to the fact that the background adversarial mixup module is model agnostic and the background-induced temporal gradient enhancement module consolidates the attack on temporal relations. This demonstrates that our attack method has promising generalization ability on the target models with different architectures.

\begin{table}[!t]
	\caption{Cross-model attack on Kinetics400.}
	\label{tbl:cross_attack_kinetics}
	\centering
	\resizebox{0.45\textwidth}{!}{%
		\begin{tabular}{lrlccc}
			\toprule[0.7pt]
			\multicolumn{1}{c}{Attack} & \multicolumn{1}{c}{Venue} & \multicolumn{1}{c}{Model} & VideoMamba & InternVideo2 & InternVideo2.5 \\ \midrule[0.5pt]
			\multirow{3}{*}{TT} & \multirow{3}{*}{AAAI'22} & NL & { 23.25} & 30.75 & { 29.00} \\
			&  & SlowFast & 19.75 & 31.00 & 24.25 \\
			&  & TPN & 16.00 & {34.25} & 25.50 \\ \midrule[0.5pt]
			ENS-I2V-MF & TPAMI'24 & Ensemble & 17.50 & 32.50 & 23.00 \\ \midrule[0.5pt]
			AENS-I2V-MF & TPAMI'24 & Ensemble & 21.75 & 33.00 & 25.50 \\ \midrule[0.5pt]
			\multirow{3}{*}{Ours} & \multicolumn{1}{l}{} & NL & \textbf{37.50} & 43.25 & \textbf{36.75} \\
			& \multicolumn{1}{l}{} & SlowFast & 28.25 & \textbf{46.00} & 33.50 \\
			& \multicolumn{1}{l}{} & TPN & 26.50 & 42.75 & 32.75 \\ \midrule[0.5pt]
		\end{tabular}%
	}
\end{table}

\begin{figure}[!t]
	\centering
	\includegraphics[width=0.145\textwidth]{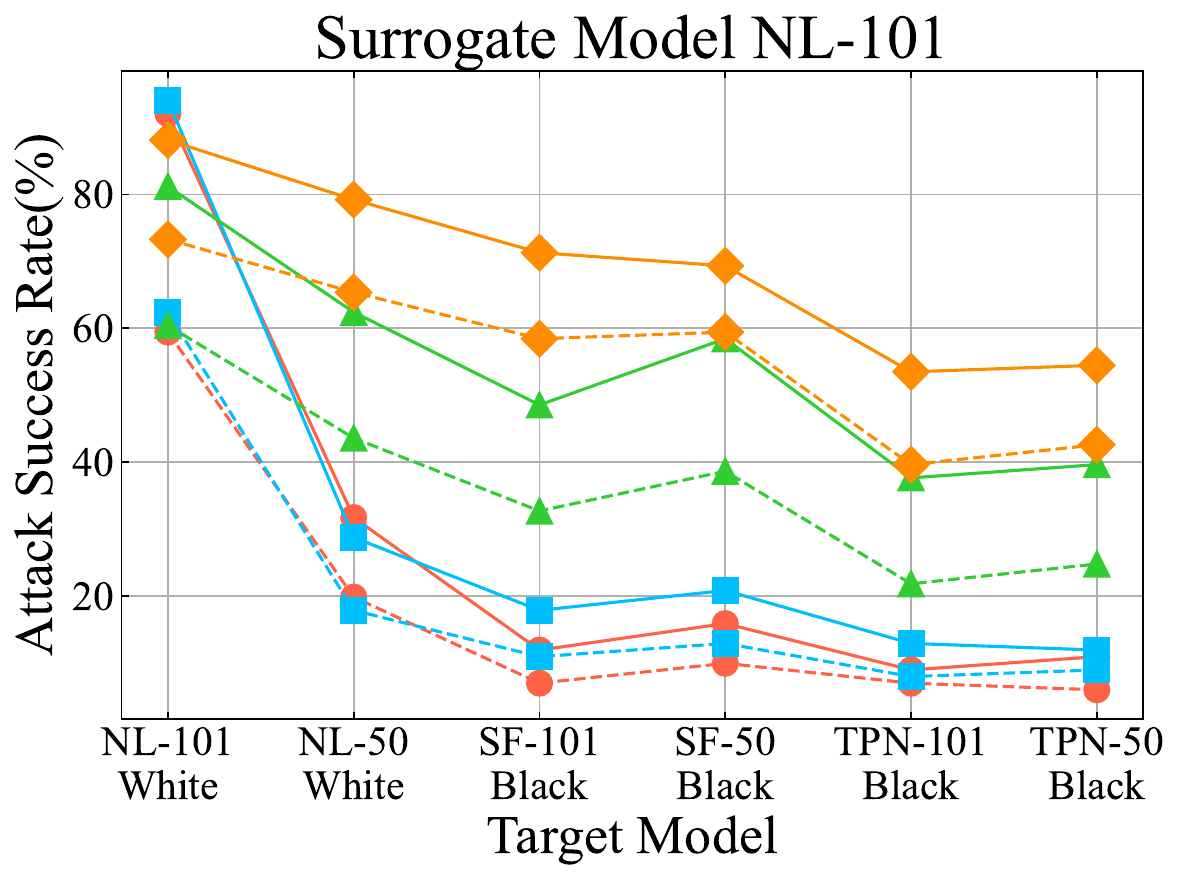}
	\includegraphics[width=0.145\textwidth]{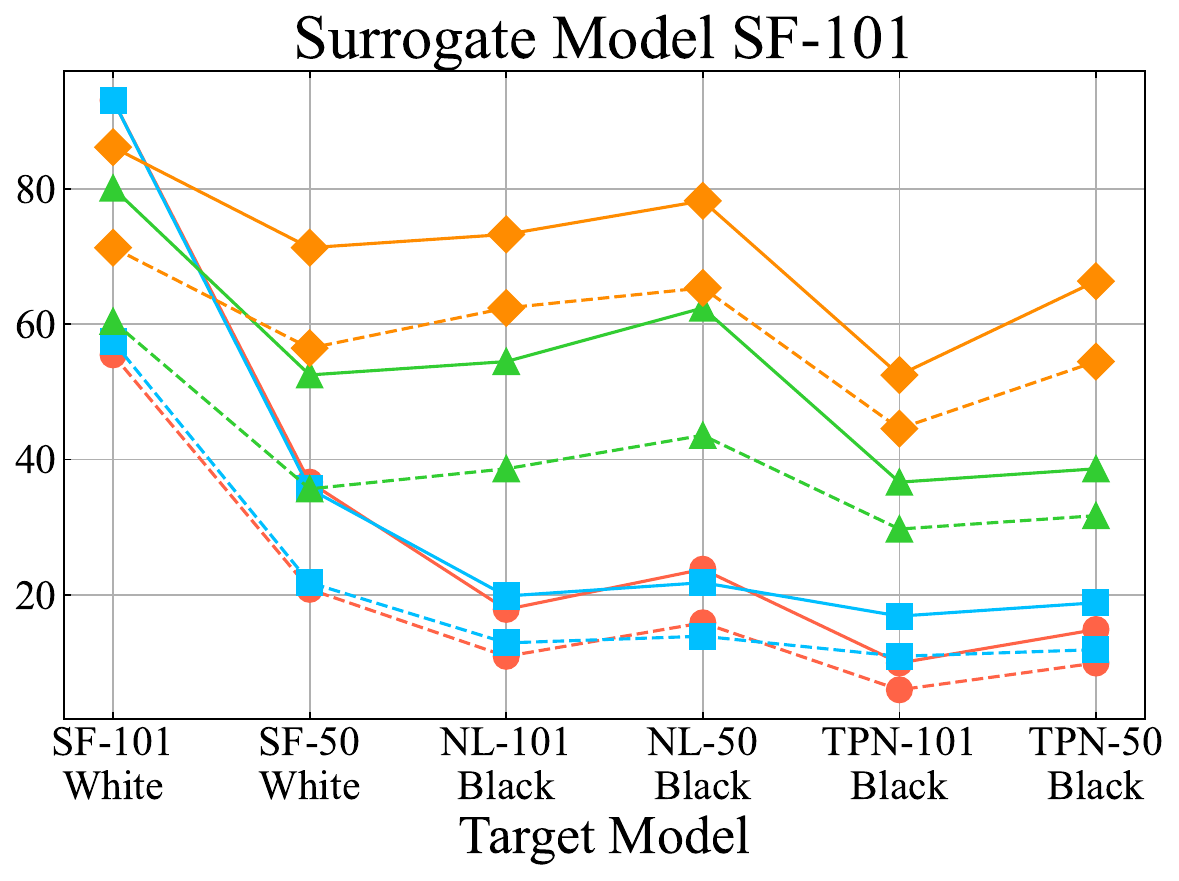}
	\includegraphics[width=0.18\textwidth]{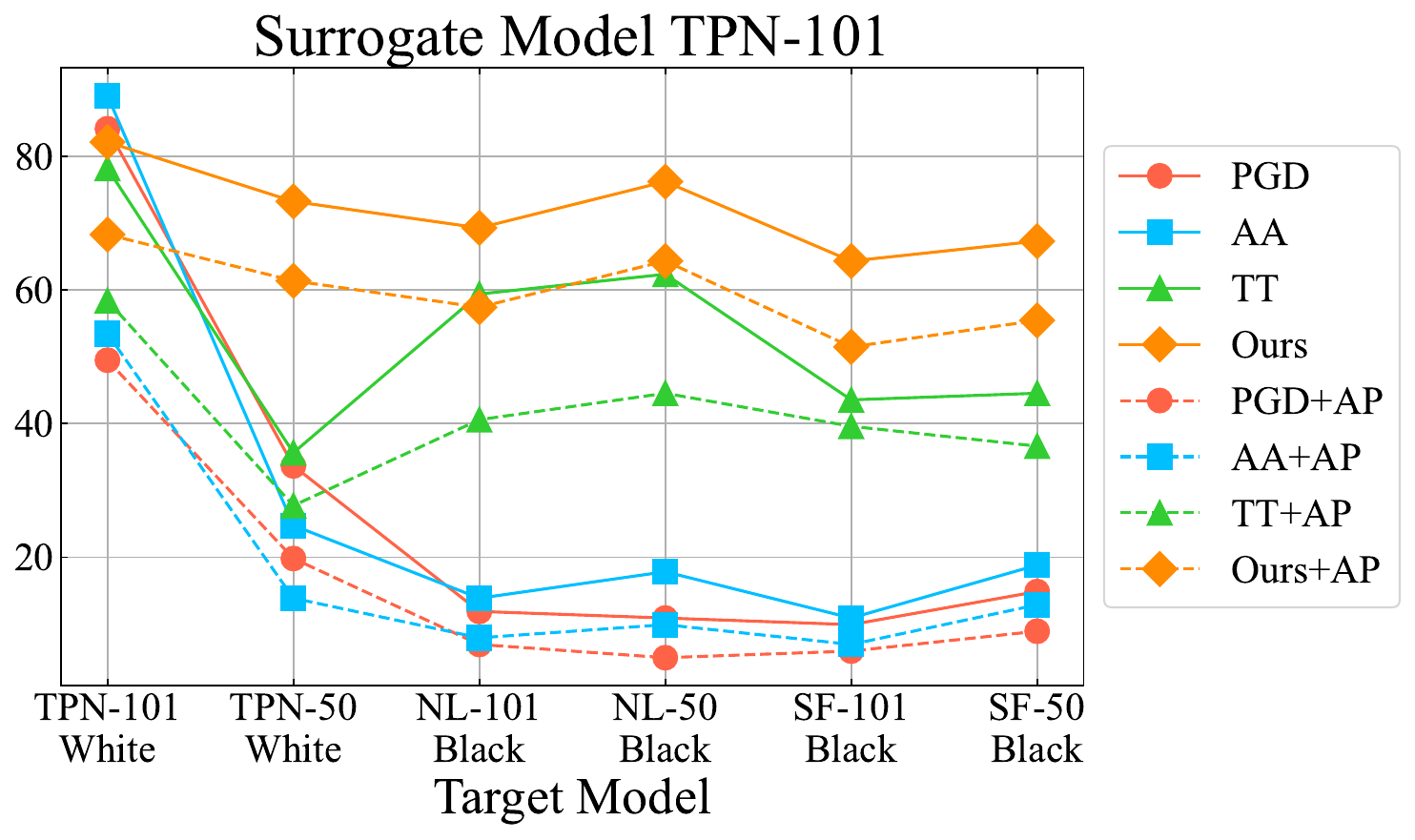}
	\caption{Transfer attack with adversarial purification on UCF101}
	\label{fig:purification_ucf}
\end{figure}

\subsection{Transfer-based attack with adversarial purification on UCF101}
For image classification, adversarial training is the most popular defense method, but it requires large computational overheads. Here we examine the performance of transfer-based attack with adversarial purification. In particular, we employ the diffusion-based adversarial purification method \cite{lee-cvpr2023-robust}, which adopts the diffusion model to handle the adversarial perturbations and it is a plug-and-play technique. The results on UCF101 are shown in Figure~\ref{fig:purification_ucf}, where the solid lines show the results without defense and the dashed lines show the results with defense, and in Table~\ref{tbl:purification_ucf}. From the results, it can be observed that our method exhibits better transfer attack ability when attacking the action recognition model with adversarial defense mechanism. Meanwhile, the adversarial purification strategy has good performance when defending white-box attack, but performs worse when being attacked by transfer-based attack, especially for the attack using input transformation. 

\begin{table}[!t]
	\centering
	\caption{Transfer-based attack with adversarial purification on UCF101.}
	\label{tbl:purification_ucf}
	\resizebox{0.5\textwidth}{!}{%
		\begin{tabular}{llrcccccc}
			\toprule[0.75pt]
			Target Models & \multicolumn{1}{c}{Attack} & \multicolumn{1}{c}{Venue} & NL-101 & NL-50 & SF-101 & SF-50 & TPN-101 & TPN-50 \\ \midrule[0.5pt]
			\multirow{4}{*}{NL-101} & PGD & ICLR'18 & \cellcolor[HTML]{C0C0C0}59.41 & \cellcolor[HTML]{C0C0C0}19.80 & 6.93 & 9.90 & 6.93 & 5.94 \\
			& AA & ICML'20 & \cellcolor[HTML]{C0C0C0}62.38 & \cellcolor[HTML]{C0C0C0}17.82 & 10.89 & 12.87 & 7.92 & 8.91 \\
			& TT & AAAI'22 & \cellcolor[HTML]{C0C0C0}60.40 & \cellcolor[HTML]{C0C0C0}43.56 & 32.67 & 38.61 & 21.78 & 24.75 \\
			& Ours &  & \cellcolor[HTML]{C0C0C0}73.26 & \cellcolor[HTML]{C0C0C0}65.35 & \textbf{58.42} & \textbf{59.41} & {\ul 39.60} & {\ul 42.57} \\ \midrule[0.5pt]
			\multirow{4}{*}{SF-101} & PGD & ICLR'18 & 10.89 & 15.84 & \cellcolor[HTML]{C0C0C0}55.45 & \cellcolor[HTML]{C0C0C0}20.79 & 5.94 & 9.90 \\
			& AA & ICML'20 & 12.87 & 13.86 & \cellcolor[HTML]{C0C0C0}57.43 & \cellcolor[HTML]{C0C0C0}21.78 & 10.89 & 11.88 \\
			& TT & AAAI'22 & 38.61 & 43.56 & \cellcolor[HTML]{C0C0C0}60.40 & \cellcolor[HTML]{C0C0C0}35.64 & 29.70 & 31.68 \\
			& Ours &  & \textbf{62.38} & \textbf{65.35} & \cellcolor[HTML]{C0C0C0}71.29 & \cellcolor[HTML]{C0C0C0}56.44 & \textbf{44.55} & \textbf{54.46} \\ \midrule[0.5pt]
			\multirow{4}{*}{TPN-101} & PGD & ICLR'18 & 6.93 & 4.95 & 5.94 & 8.91 & \cellcolor[HTML]{C0C0C0}49.50 & \cellcolor[HTML]{C0C0C0}19.80 \\
			& AA & ICML'20 & 7.92 & 9.90 & 6.93 & 12.87 & \cellcolor[HTML]{C0C0C0}53.47 & \cellcolor[HTML]{C0C0C0}13.86 \\
			& TT & AAAI'22 & 40.59 & 44.55 & 39.60 & 36.63 & \cellcolor[HTML]{C0C0C0}58.42 & \cellcolor[HTML]{C0C0C0}27.72 \\
			& Ours &  & {57.43} & {64.36} & {51.49} & {55.45} & \cellcolor[HTML]{C0C0C0}68.32 & \cellcolor[HTML]{C0C0C0}61.39 \\ \toprule[0.75pt]
		\end{tabular}%
	}
\end{table}

\section*{Acknowledgments}
This work was supported in part by Zhejiang Provincial Natural Science Foundation of China under Grant LR23F020002, and in part by Hangzhou Key Research and Development Program under Grant 2024SZD1A12. 

\small
\bibliographystyle{named}

\end{document}